\documentclass{article}

    \usepackage[final,nonatbib]{neurips_2020}

\usepackage[utf8]{inputenc} %
\usepackage[T1]{fontenc}    %
\usepackage{hyperref}       %
\usepackage{url}            %
\usepackage{booktabs}       %
\usepackage{amsfonts}       %
\usepackage{amsmath}
\usepackage{amssymb}
\usepackage{amsthm}         %
\usepackage{nicefrac}       %
\usepackage{microtype}      %
\usepackage[numbers]{natbib}
\usepackage[dvipsnames]{xcolor}
\usepackage{graphicx, wrapfig}
\usepackage{stackengine}
\usepackage{mathtools}

\usepackage[noend]{algpseudocode}
\usepackage{algorithm}

\usepackage{hyperref}
\hypersetup{
 colorlinks=True,
 linkcolor=blue,
 citecolor=blue,
 urlcolor=blue}

\newtheorem{theorem}{Theorem}[section]
\newtheorem*{theorem*}{Theorem}
\pdfoutput=1

\newtheorem*{proposition*}{Proposition}
\newtheorem{lemma}[theorem]{Lemma}
\newtheorem*{lemma*}{Lemma}

\newtheorem*{conjecture*}{Conjecture}

\newtheorem*{fact*}{Fact}

\newtheorem*{hypothesis*}{Hypothesis}

\theoremstyle{definition}

\newtheorem{assumption}[theorem]{Assumption}
\DeclareMathOperator*{\argmax}{arg\,max}
\theoremstyle{remark}

\newcommand{\hatT}{\widehat{T}}
\newcommand{\hatM}{\widehat{M}}
\newcommand{\tT}{T}
\newcommand{\tM}{M}
\newcommand{\tilM}{\widetilde{M}}

\def\shownotes{1}  \ifnum\shownotes=1
\newcommand{\authnote}[2]{{[#1: #2]}}
\else
\newcommand{\authnote}[2]{}
\fi

\newtheorem*{remark*}{Remark}

\newtheorem*{observation*}{Observation}

\newcommand{\E}{\mathbb{E}}
\newcommand{\Esub}[1]{\underset{#1}{\E}}
\newcommand{\Ei}{\bar{\mathbb{E}}}
\newcommand{\Eisub}[1]{\underset{#1}{\Ei}}
\newcommand{\dd}{\,\mathrm{d}}

\makeatletter
\newcommand{\printfnsymbol}[1]{%
  \textsuperscript{\@fnsymbol{#1}}%
}

\title{MOPO: Model-based Offline Policy Optimization}

\author{%
 Tianhe Yu\thanks{ equal contribution. $\dagger$ equal advising. Orders randomized.}~~$^1$, Garrett Thomas\printfnsymbol{1}$^1$, Lantao Yu$^1$, Stefano Ermon$^1$, James Zou$^1$,\\
 \textbf{Sergey Levine}$^2$, \textbf{Chelsea Finn}$\dagger$$^1$, \textbf{Tengyu Ma}$\dagger$$^1$\\
 Stanford University$^1$, UC Berkeley$^2$\\
 \texttt{\{tianheyu,gwthomas\}@cs.stanford.edu} \\
}

\begin{document}

\maketitle

\begin{abstract}
  Offline reinforcement learning (RL) refers to the problem of learning policies entirely from a large batch of previously collected data. This problem setting offers the promise of utilizing such datasets to acquire policies without any costly or dangerous active exploration. However, it is also challenging, due to the distributional shift between the offline training data and those states visited by the learned policy. 
  Despite significant recent progress, the most successful prior methods are model-free and constrain the policy to the support of data, precluding generalization to unseen states. In this paper, we first observe that an existing model-based RL algorithm already produces significant gains in the offline setting compared to model-free approaches. However, standard model-based RL methods, designed for the online setting, do not provide an explicit mechanism to avoid the offline setting's distributional shift issue. Instead, we propose to modify the existing model-based RL methods by applying them with rewards artificially penalized by the uncertainty of the dynamics. We theoretically show that the algorithm maximizes a lower bound of the policy's return under the true MDP. We also characterize the trade-off between the gain and risk of leaving the support of the batch data.  Our algorithm, Model-based Offline Policy Optimization (MOPO), outperforms standard model-based RL algorithms and prior state-of-the-art model-free offline RL algorithms on existing offline RL benchmarks and two challenging continuous control tasks that require generalizing from data collected for a different task.
\end{abstract}

\section{Introduction}

Recent advances in machine learning using deep neural networks have shown significant successes in scaling
to large realistic datasets, such as ImageNet~\cite{deng2009imagenet} in computer vision, SQuAD~\cite{rajpurkar2016squad} in NLP, and RoboNet~\cite{dasari2019robonet} in robot learning. Reinforcement learning (RL) methods, in contrast, struggle to scale to many real-world applications, e.g., autonomous driving~\cite{yu2018bdd100k} and healthcare~\cite{gottesman2019guidelines}, because they rely on costly online trial-and-error. %
However, pre-recorded %
datasets in domains like these %
can be large and diverse. Hence, designing RL algorithms that can learn from those diverse, static datasets would both enable more practical RL training in the real world and lead to more effective generalization.

While off-policy RL algorithms~\cite{lillicrap2015continuous, haarnoja2018soft, fujimoto2018addressing} can in principle utilize previously collected datasets, they perform poorly without online data collection. 
These failures are generally caused by %
large extrapolation error when the Q-function is evaluated on out-of-distribution actions~\cite{fujimoto2018off, kumar2019stabilizing}, which can lead to unstable learning and divergence.
Offline RL methods propose to mitigate bootstrapped error by constraining the learned policy to the behavior policy induced by the dataset~\cite{fujimoto2018off,kumar2019stabilizing,wu2019behavior,jaques2019way,nachum2019algaedice,peng2019advantage,siegel2020keep}. While these methods achieve reasonable performances in some settings, their learning is limited to behaviors within the data manifold. Specifically, these methods estimate error
with respect to out-of-distribution \emph{actions}, but only consider \emph{states} that lie within the offline dataset and do not consider those that are out-of-distribution. %
We argue that it is important for an offline RL algorithm to be equipped with the ability to leave the data support to learn a better policy for two reasons: (1) the provided batch dataset is usually sub-optimal in terms of both the states and actions covered by the dataset, and (2) the target task can be different from the tasks performed in the batch data for various reasons, e.g., because data is not available or hard to collect for the target task.
Hence, the central question that this work is trying to answer is: can we develop an offline RL algorithm that generalizes beyond the state and action support of the offline data? %

\begin{wrapfigure}{r}{0.5\linewidth}
\vspace{-0.45cm}
\centering
\small
\includegraphics[width=0.495\textwidth]{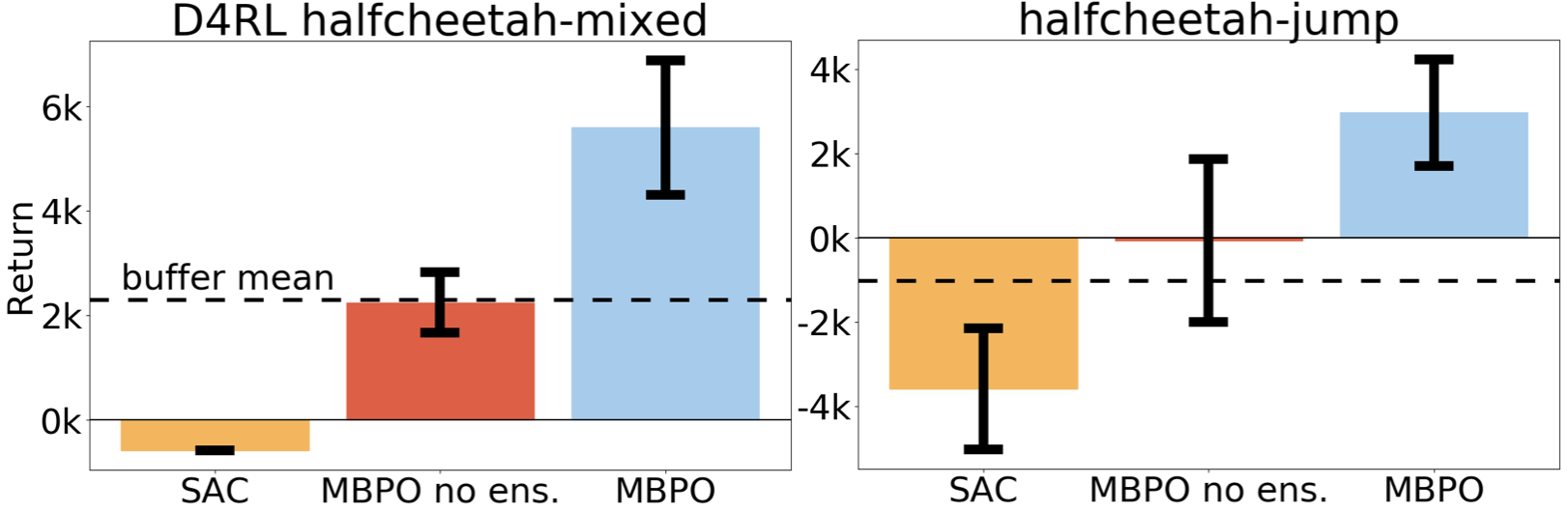}
\vspace{-0.55cm}
\caption{\footnotesize Comparison between vanilla model-based RL (MBPO~\cite{janner2019trust}) with or without model ensembles and vanilla model-free RL (SAC~\cite{haarnoja2018soft}) on two offline RL tasks: one from the D4RL benchmark~\cite{fu2020d4rl} and one that demands out-of-distribution generalization. We find that MBPO substantially outperforms SAC, providing some evidence that model-based approaches are well-suited for batch RL. For experiment details, see Section~\ref{sec:experiments}.
}
\label{fig:vanilla}
\normalsize
\vspace{-0.4cm}
\end{wrapfigure}
To approach this question, we first hypothesize that model-based RL methods~\cite{sutton1991dyna, deisenroth2011pilco, levine2013guided, kumar2016optimal, janner2019trust, luo2018algorithmic} make a natural choice for enabling %
generalization, for a number of reasons. First, model-based RL algorithms effectively receive more supervision, since the model is trained on every transition, even in sparse-reward settings.
Second, they are trained with supervised learning, which provides more stable and less noisy gradients than bootstrapping.
Lastly, uncertainty estimation techniques, such as bootstrap ensembles, are well developed for supervised learning methods~\cite{lakshminarayanan2017simple,kuleshov2018accurate,snoek2019can} %
and are known to perform poorly for value-based RL methods~\cite{wu2019behavior}. %
All of these attributes have the potential to improve or control generalization. %
As a proof-of-concept experiment, we evaluate two state-of-the-art off-policy
model-based and model-free algorithms, MBPO~\cite{janner2019trust} and SAC~\citep{haarnoja2018soft}, in Figure~\ref{fig:vanilla}. 
Although neither method is designed for the batch setting, we find that the model-based method and its variant without ensembles show surprisingly large gains.  This finding corroborates our hypothesis, suggesting that model-based methods are particularly well-suited for the batch setting, motivating their use in this paper.

Despite these promising preliminary results, we expect significant headroom for improvement. In particular, because offline model-based algorithms cannot improve the dynamics model using additional experience, we expect that such algorithms require careful use of the model in regions outside of the data support.
Quantifying the risk imposed by imperfect dynamics and appropriately trading off that risk with the return is a key ingredient towards building a strong offline model-based RL algorithm. 
To do so, we modify MBPO to incorporate a \emph{reward penalty} based on an estimate of the model error. Crucially, this estimate is model-dependent, and does not necessarily penalize all out-of-distribution states and actions equally, but rather prescribes penalties based on the estimated magnitude of model error. Further, this estimation is done both on \emph{states} and \emph{actions}, allowing generalization to both, in contrast to model-free approaches that only reason about uncertainty with respect to actions.

The primary contribution of this work is an offline model-based RL algorithm that optimizes a policy in an uncertainty-penalized MDP, where the reward function is penalized by an estimate of the model's error.
Under this new MDP, we theoretically show that we maximize a lower bound of the return in the true MDP, and find the optimal trade-off between the return and the risk. 
Based on our analysis, we develop a practical method that estimates model error using the predicted variance of a learned model, uses this uncertainty estimate as a reward penalty, and trains a policy using MBPO in this uncertainty-penalized MDP.
We empirically compare this approach, model-based offline policy optimization (MOPO), to both MBPO and existing state-of-the-art model-free offline RL algorithms.
Our results suggest that MOPO substantially outperforms these prior methods on the offline RL benchmark D4RL~\cite{fu2020d4rl} as well as on offline RL problems where the agent must generalize to out-of-distribution states in order to succeed.

\section{Related Work}

Reinforcement learning algorithms are well-known for their ability to acquire behaviors through online trial-and-error in the environment~\cite{barto1983neuronlike, sutton1998reinforcement}. However, such online data collection can incur high sample complexity~\cite{mnih2016asynchronous, schulman2015trust, schulman2017proximal}, limit the power of generalization to unseen random initialization
~\cite{cobbe2018quantifying, zhang2018dissection, bengio2020interference}, and pose risks in safety-critical settings~\cite{thomas2015safe}. These requirements often make real-world applications of RL less feasible. To overcome some of these challenges, we study the batch offline RL setting~\cite{lange2012batch}. 
While many off-policy RL algorithms~\cite{precup2001off, degris2012off, jiang2015doubly, munos2016safe, lillicrap2015continuous, haarnoja2018soft, fujimoto2018addressing, gu2016q, gu2017interpolated} can in principle be applied to a batch offline setting, they perform poorly in practice~\cite{fujimoto2018off}. %

\textbf{Model-free Offline RL. }
Many model-free batch RL methods are designed with two main ingredients: (1) constraining the learned policy to be closer to the behavioral policy either explicitly~\cite{fujimoto2018off, kumar2019stabilizing, wu2019behavior, jaques2019way, nachum2019algaedice} or implicitly~\cite{peng2019advantage, siegel2020keep}, and (2) applying uncertainty quantification techniques, such as ensembles, to stabilize Q-functions~\cite{agarwal2019striving, kumar2019stabilizing, wu2019behavior}.
In contrast, our \textit{model-based} method does not rely on constraining the policy to the behavioral distribution, allowing the policy to potentially benefit from taking actions outside of it. Furthermore, we utilize uncertainty quantification to quantify the risk of leaving the behavioral distribution and trade it off with the gains of exploring diverse states. %

\textbf{Model-based Online RL.} Our approach builds upon the wealth of prior work on model-based online RL methods that model the dynamics by Gaussian processes~\cite{deisenroth2011pilco}, local linear models~\cite{levine2013guided, kumar2016optimal}, neural network function approximators~\cite{draeger1995model, gal2016improving, depeweg2016learning}, and neural video prediction models~\cite{ebert2018visual, kaiser2019modelbased}. Our work is orthogonal to the choice of model.
While prior approaches have used these models to select actions using planning~\cite{tamar2016value, finn2017deep, racaniere2017imagination, oh2017value, silver2017predictron, wang2019exploring}, we choose to build upon Dyna-style approaches that optimize for a policy~\cite{sutton1991dyna, sutton2012dyna, yao2009multi, kaiser2019modelbased, ha2018world, holland2018effect,luo2018algorithmic},  specifically MBPO~\cite{janner2019trust}.
See~\cite{wang2019benchmarking} for an empirical evaluation of several model-based RL algorithms.
Uncertainty quantification, a key ingredient to our approach, is critical to good performance in model-based RL both theoretically~\citep{strehl2008analysis, zanettetighter, luo2018algorithmic} and empirically~\cite{deisenroth2011pilco, chua2018deep, nagabandi2019deep, kurutach2018model, clavera2018model}, and in optimal control~\cite{ stengel1994optimal, banaszuk2011scalable, kim2013wiener}.
Unlike these works, we develop and leverage proper uncertainty estimates that particularly suit the offline setting. 

Concurrent work by \citet{kidambi2020morel} also develops an offline model-based RL algorithm, MOReL. Unlike MOReL, which constructs terminating states based on a hard threshold on uncertainty, MOPO uses a soft reward penalty to incorporate uncertainty. In principle, a potential benefit of a soft penalty is that the policy is allowed to take a few risky actions and then return to the confident area near the behavioral distribution without being terminated. 
Moreover, while~\citet{kidambi2020morel} compares to model-free approaches, we make the further observation that even a vanilla model-based RL method outperforms model-free ones in the offline setting, opening interesting questions for future investigation. Finally, we evaluate our approach on both standard benchmarks~\cite{fu2020d4rl} and domains that require out-of-distribution generalization, achieving positive results in both. %

\newcommand{\D}{{\mathcal{D}}}
\newcommand{\Denv}{{\mathcal{D}_\text{env}}}
\newcommand{\Dmodel}{{\mathcal{D}_\text{model}}}
\newcommand{\given}{\,|\,}
\newcommand{\pib}{\pi^{\textup{B}}}

\section{Preliminaries}

We consider the standard Markov decision process (MDP) $M = (\mathcal{S}, \mathcal{A}, \tT, r, \mu_0, \gamma)$, where $\mathcal{S}$ and $\mathcal{A}$ denote the state space and action space respectively, $\tT(s' \given s, a)$  the transition dynamics, $r(s, a)$  the reward function, $\mu_0$  the initial state distribution, and $\gamma \in (0, 1)$  the discount factor. The goal in RL is to optimize a policy $\pi(a \given s)$ that maximizes the expected discounted return $\eta_{\tM}(\pi) := \Esub{\pi, \tT, \mu_0}\left[\sum_{t=0}^\infty \gamma^t r(s_t, a_t)\right]$. The value function $V^\pi_{\tM}(s) := \Esub{\pi, \tT}\left[\sum_{t=0}^\infty \gamma^t r(s_t, a_t) \given s_0=s\right]$ gives the expected discounted return under $\pi$ when starting from state $s$.

In the \textit{offline RL} problem, the algorithm only has access to a static dataset $\D_{\textup{env}} = \{(s, a, r, s')\}$ collected by one or a mixture of behavior policies $\pib$, and cannot interact further with the environment. We refer to the distribution from which $\D_{\textup{env}}$ was sampled as the \textit{behavioral distribution}.

We also introduce the following notation for the derivation in Section~\ref{sec:mopo}.
In the model-based approach we will have a dynamics model $\hatT$ estimated from the transitions in $\D_{\textup{env}}$.
This \textit{estimated dynamics} defines a \textit{model MDP} $\hatM = (\mathcal{S}, \mathcal{A}, \hatT, r, \mu_0, \gamma)$.
Let $\mathbb{P}^\pi_{\hatT,t}(s)$ denote the probability of being in state $s$ at time step $t$ if actions are sampled according to $\pi$ and transitions according to $\hatT$. Let $\rho^\pi_{\hatT}(s,a)$ be the discounted occupancy measure of policy $\pi$ under dynamics $\hatT$:
$\rho^\pi_{\hatT}(s,a) := \pi(a \given s) \sum_{t=0}^\infty \gamma^t \mathbb{P}^\pi_{\hatT,t}(s)$. Note that $\rho^\pi_{\hatT}$, as defined here, is not a properly normalized probability distribution, as it integrates to $1/(1-\gamma)$. We will denote (improper) expectations with respect to $\rho^\pi_{\hatT}$ with $\Ei$, as in $\eta_{\hatM}(\pi) = \Ei_{\rho^\pi_{\hatT}}[r(s,a)]$.

We now summarize model-based policy optimization (MBPO)~\cite{janner2019trust}, which we build on in this work.  MBPO learns a model of the transition distribution $\hatT_\theta(s' \vert s, a)$ parametrized by $\theta$, via supervised learning on the behavorial data $\Denv$. MBPO also learns a model of the reward function in the same manner. During training, MBPO performs $k$-step rollouts using $\hatT_\theta(s' \vert s, a)$ starting from state $s \in \Denv$, adds the generated data to a separate replay buffer $\Dmodel$, and finally updates the policy $\pi(a \vert s)$ using data sampled from $\Denv \cup \Dmodel$.
When applied in an online setting, MBPO iteratively collects samples from the environment and uses them to further improve both the model and the policy.
In our experiments in Table~\ref{fig:d4rl}, Table~\ref{sec:ood} and Table~\ref{fig:vanilla}, we observe that MBPO performs surprisingly well on the offline RL problem compared to model-free methods. In the next section, we derive MOPO, which builds upon MBPO to further improve performance.

\section{MOPO: Model-Based Offline Policy Optimization}
\label{sec:mopo}

Unlike model-free methods, our goal is to design an offline model-based reinforcement learning algorithm that can take actions that are not strictly within the support of the behavioral distribution. Using a model gives us the potential to do so. However, models will become increasingly inaccurate further from the behavioral distribution, and vanilla model-based policy optimization algorithms may exploit these regions where the model is inaccurate. This concern is especially important in the offline setting, where mistakes in the dynamics will not be corrected with additional data collection.

For the algorithm to perform reliably, it's crucial to balance the return and risk: 1. the potential gain in performance by escaping the behavioral distribution and finding a better policy, and 2. the risk of overfitting to the errors of the dynamics at regions far away from the behavioral distribution.  To achieve the optimal balance, we first bound the return from below by the return of a constructed model MDP penalized by the uncertainty of the dynamics (Section~\ref{sec:uq}). Then we maximize the conservative estimation of the return by an off-the-shelf reinforcement learning algorithm, which gives MOPO, a generic model-based off-policy algorithm (Section~\ref{sec:alg}). We discuss important practical implementation details in Section~\ref{sec:imp}.

\subsection{Quantifying the uncertainty: from the dynamics to the total return}\label{sec:uq}

Our key idea is to build a lower bound for the expected return of a policy $\pi$ under the true dynamics and then maximize the lower bound over $\pi$.
A natural estimator for the true return $\eta_{\tM}(\pi)$ is $\eta_{\hatM}(\pi)$, the return under the estimated dynamics.
The error of this estimator depends on, potentially in a complex fashion, the error of $\hatM$, which may compound over time.  In this subsection, we characterize how the error
of $\hatM$ influences the uncertainty of the total return. We begin by stating a lemma (adapted from \cite{luo2018algorithmic}) that gives a precise relationship between the performance of a policy under dynamics $T$ and dynamics $\hatT$. (All proofs are given in Appendix~\ref{app:proofs}.)

\begin{lemma}[Telescoping lemma] Let $M $ and $\hatM$ be two MDPs with the same reward function $r$, but different dynamics $\tT$ and $\hatT$ respectively. 
Let $G^\pi_{\hatM}(s,a) := \Esub{s' \sim \hatT(s,a)}[V_M^\pi(s')] - \Esub{s' \sim \tT(s,a)}[V_M^\pi(s')]$.
Then,
\begin{align}
\eta_{\hatM}(\pi) - \eta_{\tM}(\pi) = \gamma \Eisub{(s,a) \sim \rho^\pi_{\hatT}}\left[G^\pi_{\hatM}(s,a)\right]\label{eqn:12}
\end{align}
\end{lemma}
\label{lem:1}

As an immediate corollary, we have
\begin{align}
\eta_{\tM}(\pi) = \Eisub{(s,a) \sim \rho^\pi_{\hatT}}\left[r(s,a) - \gamma G^\pi_{\hatM}(s,a)\right] \ge \Eisub{(s,a) \sim \rho^\pi_{\hatT}}\left[r(s,a) - \gamma |G^\pi_{\hatM}(s,a)|\right] \label{eqn:1}
\end{align}

Here and throughout the paper, we view $\tT$ as the real dynamics and $\hatT$ as the learned dynamics. We observe that the  quantity $G^\pi_{\hatM}(s,a)$ plays a key role linking the estimation error of the dynamics and the estimation error of the return. 
By definition, we have that $G^\pi_{\hatM}(s,a)$ measures the difference between $\tM$ and $\hatM$ under the test function $V^\pi$ --- indeed, if $M = \hatM$, then $G^\pi_{\hatM}(s,a) = 0$.
By equation~\eqref{eqn:12}, it governs the differences between the performances of $\pi$ in the two MDPs.  
If we could estimate $G^\pi_{\hatM}(s,a)$ or bound it from above, then we could use the RHS of ~\eqref{eqn:12} as an upper bound for the estimation error of $\eta_{\tM}(\pi)$. Moreover, 
equation~\eqref{eqn:1} suggests that a policy that obtains high reward in the estimated MDP while also minimizing $G^\pi_{\hatM}$ will obtain high reward in the real MDP. 

However, computing $G^\pi_{\hatM}$ remains elusive because it depends on the unknown function $V_{\tM}^\pi$. 
Leveraging properties of $V_M^\pi$, we will replace $G^\pi_{\hatM}$ by an upper bound that depends solely on the error of the dynamics  $\hatT$. 
We first note that if $\mathcal{F}$ is a set of functions  mapping $\mathcal{S}$ to $\mathbb{R}$ that contains $V^\pi_{\tM} $, then,
\begin{align}
|G^\pi_{\hatM}(s,a)| \le \sup_{f \in \mathcal{F}} \left|\Esub{s' \sim \hatT(s,a)} [f(s')] - \Esub{s' \sim \tT(s,a)}[f(s')]\right| =: d_{\mathcal{F}}(\hatT(s,a), \tT(s,a)), \label{eqn:13}
\end{align}
where $d_{\mathcal{F}}$ is the integral probability metric (IPM)~\cite{muller1997integral} defined by $\mathcal{F}$.
IPMs are quite general and contain several other distance measures as special cases~\cite{sriperumbudur2009integral}.
Depending on what we are willing to assume about $V^\pi_{\tM}$, there are multiple options to bound $G^\pi_{\hatM}$ by some notion of error of $\hatT$, discussed in greater detail in Appendix~\ref{app:ipms}:

(i) If $\mathcal{F} = \{f : \|f\|_\infty \le 1\}$, then $d_{\mathcal{F}}$ is the \textit{total variation distance}.
Thus, if we assume that the reward function is bounded such that $\forall (s,a), ~ |r(s,a)| \le r_{\max}$, we have
$\|V^\pi\|_\infty \le \sum_{t=0}^\infty \gamma^t r_{\max} = \frac{r_{\max}}{1-\gamma}$, and hence
\begin{align}
|G^\pi_{\hatM}(s,a)| \le \frac{r_{\max}}{1-\gamma} D_{\textsc{tv}}(\hatT(s,a), \tT(s,a))
\end{align}

(ii) If $\mathcal{F}$ is the set of 1-Lipschitz function w.r.t. to some distance metric, then $d_\mathcal{F}$ is the \textit{1-Wasserstein distance} w.r.t. the same metric.
Thus, if we assume that $V^\pi_{\tM}$ is $L_v$-Lipschitz with respect to a norm $\|\cdot\|$, it follows that
\begin{align}
|G^\pi_{\hatM}(s,a)| & \le L_v W_1(\hatT(s,a), \tT(s,a))%
\end{align}
Note that when $\hatT$ and $\tT$ are both deterministic, then $W_1(\hatT(s,a), \tT(s,a)) =  \|\hatT(s,a) - \tT(s,a)\|$ (here $\tT(s,a)$ denotes the deterministic output of the model $\tT$). 

Approach (ii) has the advantage that it incorporates the geometry of the state space, but at the cost of an additional assumption which is generally impossible to verify in our setting. The assumption in (i), on the other hand, is extremely mild and typically holds in practice. Therefore we will prefer (i) unless we have some prior knowledge about the MDP. We summarize the assumptions and the inequalities in the options above as follows.

\begin{assumption}\label{ass:1}
	Assume a scalar $c$ and a function class $\mathcal{F}$ such that $V^\pi_{\tM}\in c\mathcal{F}$ for all $\pi$.
	\end{assumption}

As a direct corollary of Assumption~\ref{ass:1} and equation~\eqref{eqn:13}, we have 
\begin{align}
|G^\pi_{\hatM}(s,a)| \le cd_{\mathcal{F}}(\hatT(s,a), \tT(s,a)).\label{eqn:16}
\end{align}

Concretely, option (i) above corresponds to $c = r_{\max}/(1-\gamma)$ and $\mathcal{F} = \{f : \|f\|_\infty \le 1\}$, and option (ii) corresponds to $c = L_v$ and $\mathcal{F} = \{f : \text{$f$ is 1-Lipschitz}\}$. We will analyze our framework under the assumption that we have access to an oracle uncertainty quantification module that provides an upper bound on the error of the model. In our implementation, we will estimate the error of the dynamics by heuristics (see sections \ref{sec:imp} and \ref{app:ablation}). %

\begin{assumption}\label{ass:2} Let $\mathcal{F}$ be the function class in Assumption~\ref{ass:1}.
	We say $u : \mathcal{S} \times \mathcal{A} \to \mathbb{R}$ is an admissible error estimator for $\widehat{T}$ if $d_{\mathcal{F}}(\hatT(s,a), \tT(s,a)) \le u(s,a)$ for all $s \in \mathcal{S}, a \in \mathcal{A}$.\footnote{The definition here extends the definition of admissible confidence interval in~\citep{strehl2008analysis} slightly to the setting of stochastic dynamics.}
\end{assumption}

Given an admissible error estimator, we define the \textit{uncertainty-penalized reward} $\tilde{r}(s,a) := r(s,a) - \lambda u(s,a)$ where $\lambda := \gamma c$, and the \textit{uncertainty-penalized MDP} $\widetilde{M} = (\mathcal{S}, \mathcal{A}, \hatT, \tilde{r}, \mu_0, \gamma)$. We observe that $\tilM$ is conservative in that the return under it bounds from below the true return:
\begin{align}
\eta_{\tM}(\pi) &\ge \Eisub{(s,a) \sim \rho^\pi_{\hatT}}\left[r(s,a) - \gamma |G^\pi_{\hatM}(s,a)|\right] \ge \Eisub{(s,a) \sim \rho^\pi_{\hatT}}\left[r(s,a) - \lambda u(s,a)\right]\tag{by equation~\eqref{eqn:1} and ~\eqref{eqn:16}} \nonumber \\
& \ge \Eisub{(s,a) \sim \rho^\pi_{\hatT}}\left[\tilde{r}(s,a)\right]  = \eta_{\tilM}(\pi)\label{eqn:5} 
\end{align} 

\subsection{Policy optimization on uncertainty-penalized MDPs}\label{sec:alg}

Motivated by ~\eqref{eqn:5}, we optimize the policy on the uncertainty-penalized MDP $\tilM$ in Algorithm~\ref{alg:1}.

\begin{algorithm}[t]

\caption{Framework for Model-based Offline Policy Optimization (MOPO) with Reward Penalty
}
\label{alg:1}
	\begin{algorithmic}[1]
		\Require Dynamics model $\widehat{T}$ with admissible error estimator $u(s,a)$; constant $\lambda$. 
		\State Define $\tilde{r}(s,a) = r(s,a) - \lambda u(s,a)$. Let $\widetilde{M}$ be the MDP with dynamics $\widehat{T}$ and reward $\tilde{r}$. 
		\State Run any RL algorithm on $\widetilde{M}$ until convergence to obtain 
		$\hat{\pi} = \textup{argmax}_{\pi} \eta_{\widetilde{M}}(\pi)$
	\end{algorithmic}
\end{algorithm}

\textbf{Theoretical Guarantees for MOPO.} We will theoretical analyze the algorithm by establishing the optimality of the learned policy $\hat{\pi}$ among a family of policies. Let $\pi^\star$ be the optimal policy on $\tM$ and $\pib$ be the policy that generates the batch data.
Define $\epsilon_u(\pi)$ as
\begin{align}
\epsilon_u(\pi) := \Eisub{(s,a)\sim \rho^\pi_{\hatT}}[u(s,a)]\label{eqn:11}
\end{align}
Note that $\epsilon_u$ depends on $\hatT$, but we omit this dependence in the notation for simplicity.
We observe that $\epsilon_u(\pi)$ characterizes how erroneous the model is along trajectories induced by $\pi$.
For example, consider the extreme case when $\pi = \pib$. 
Because $\hatT$ is learned on the data generated from $\pib$, we expect $\hatT$ to be relatively accurate for those $(s,a)\sim \rho^{\pib}_{\hatT}$, and thus $u(s,a)$ tends to be small. Thus, we expect $\epsilon_u(\pib)$ to be quite small. On the other end of the spectrum, 
when $\pi$ often visits states out of the batch data distribution in the real MDP, namely $\rho^{\pi}_T$ is different from $\rho^{\pib}_{\tT}$, we expect that $\rho^{\pi}_{\hatT}$ is even more different from the batch data and therefore the error estimates $u(s,a)$ for those $(s,a)\sim \rho^\pi_{\hatT}$ tend to be large. As a consequence, we have that $\epsilon_u(\pi)$ will be large. 

For $\delta \ge \delta_{\min} := \min_\pi \epsilon_u(\pi)$, let $\pi^{\delta}$ be the best policy among those incurring model error at most $\delta$:
\begin{align}
\pi^\delta := \argmax_{\pi: \epsilon_u(\pi)\le \delta} ~\eta_{\tM}(\pi) \label{eqn:7}
\end{align}

The main theorem provides a performance guarantee on the policy $\hat\pi$ produced by MOPO.
\begin{theorem} Under Assumption~\ref{ass:1} and \ref{ass:2}, the learned policy $\hat{\pi}$ in MOPO (Algorithm~\ref{alg:1}) satisfies
	\begin{align}
			\eta_{\tM}(\hat{\pi}) \ge \sup_{\pi}\{\eta_{\tM}(\pi) - 2\lambda\epsilon_u(\pi)\} \label{eqn:14}
	\end{align}
	In particular, for all $\delta \ge \delta_{\min}$,
	\begin{align}
	    \hspace{1cm} \eta_{\tM}(\hat{\pi}) \ge \eta_{\tM}(\pi^\delta) - 2\lambda\delta \label{eqn:15}
	\end{align}
\end{theorem}
\label{thm:main}

\noindent{\bf Interpretation:} One consequence of~\eqref{eqn:14} is that $\eta_{\tM}(\hat\pi) \ge \eta_{\tM}(\pib) - 2\lambda \epsilon_u(\pib)$.
This suggests that $\hat{\pi}$ should perform at least as well as the behavior policy $\pib$, because, as argued before, $\epsilon_u(\pib)$ is expected to be small.

Equation~\eqref{eqn:15} tells us that the learned policy $\hat{\pi}$ can be as good as any policy $\pi$ with $\epsilon_u(\pi)\le \delta$, or in other words, any policy that visits states with sufficiently small uncertainty as measured by $u(s,a)$.
A special case of note is when $\delta = \epsilon_u(\pi^\star)$, we have $\eta_{\tM}(\hat\pi) \ge \eta_{\tM}(\pi^\star) - 2\lambda\epsilon_u(\pi^\star)$, which suggests that the suboptimality gap between the learned policy $\hat{\pi}$ and the optimal policy $\pi^\star$ depends on the error $\epsilon_u(\pi^\star)$. 
The closer $\rho^{\pi^\star}_{\hatT}$ is to the batch data, the more likely the uncertainty $u(s,a)$ will be smaller on those points $(s,a)\sim \rho^{\pi^\star}_{\hatT}$. On the other hand, the smaller the uncertainty error of the dynamics is, the smaller $\epsilon_u(\pi^\star)$ is. In the extreme case when $u(s,a) = 0$ (perfect dynamics and uncertainty quantification), we recover the optimal policy $\pi^\star$.

Second, by varying the choice of $\delta$ to maximize the RHS of Equation~\eqref{eqn:15}, we trade off the risk and the return.  As $\delta$ increases, the return $\eta_{\tM}(\pi^\delta)$ increases also, since $\pi^\delta$ can be selected from a larger set of policies. However, the risk factor $2\lambda\delta$ increases also. The optimal choice of $\delta$ is achieved when the risk balances the gain from exploring policies far from the behavioral distribution. The exact optimal choice of $\delta$ may depend on the particular problem.   We note $\delta$ is only used in the analysis, and our algorithm \textit{automatically achieves the optimal balance} because Equation~\eqref{eqn:15} holds for any $\delta$.
\subsection{Practical implementation}\label{sec:imp}

Now we describe a practical implementation of MOPO motivated by the analysis above.
The method is summarized in Algorithm~\ref{alg:practical} in Appendix~\ref{app:alg}, and largely follows MBPO with a few key exceptions.

Following MBPO, we model the dynamics using a neural network that outputs a Gaussian distribution over the next state and reward\footnote{If the reward function is known, we do not have to estimate the reward. The theory in Sections \ref{sec:uq} and \ref{sec:alg} applies to the case where the reward function is known. To extend the theory to an unknown reward function, we can consider the reward as being concatenated onto the state, so that the admissible error estimator bounds the error on $(s',r)$, rather than just $s'$.}: $\hatT_{\theta, \phi}(s_{t+1}, r|s_t, a_t) = \mathcal{N}(\mu_\theta(s_t, a_t), \Sigma_\phi(s_t, a_t))$. We learn an ensemble of $N$ dynamics models $\{\hatT^i_{\theta, \phi} = \mathcal{N}(\mu^i_\theta, \Sigma^i_\phi)\}_{i=1}^N$, with each model trained independently via maximum likelihood. 

The most important distinction from MBPO is that we use uncertainty quantification following the analysis above. We aim to design the uncertainty estimator that captures both the epistemic and aleatoric uncertainty of the true dynamics. Bootstrap ensembles have been shown to give a consistent estimate of the population mean in theory~\cite{bickel1981some} and empirically perform well in model-based RL~\cite{chua2018deep}. Meanwhile, the learned variance of a Gaussian probabilistic model can theoretically recover the true aleatoric uncertainty when the model is well-specified. To leverage both, we design our error estimator $u(s,a) = \max_{i=1}^N\|\Sigma^i_\phi(s, a)\|_\text{F}$, the maximum standard deviation of the learned models in the ensemble. We use the maximum of the ensemble elements rather than the mean to be more conservative and robust. While this estimator lacks theoretical guarantees, we find that it is sufficiently accurate to achieve good performance in practice.\footnote{Designing prediction confidence intervals with strong theoretical guarantees is challenging and beyond the scope of this work, which focuses on using uncertainty quantification properly in offline RL.} Hence the practical uncertainty-penalized reward of MOPO is computed as 
$\tilde{r}(s,a) = \hat{r}(s,a) - \lambda \max_{i=1,\dots,N}\|\Sigma^i_\phi(s,a)\|_\text{F}$
where $\hat{r}$ is the mean of the predicted reward output by $\hatT$.

We treat the penalty coefficient $\lambda$ as a user-chosen hyperparameter.
Since we do not have a true admissible error estimator, the value of $\lambda$ prescribed by the theory may not be an optimal choice in practice; it should be larger if our heuristic $u(s,a)$ underestimates the true error and smaller if $u$ substantially overestimates the true error.

\section{Experiments}
\label{sec:experiments}

In our experiments, we aim to study the follow questions: (1) How does MOPO perform on standard offline RL benchmarks in comparison to prior state-of-the-art approaches?
(2) Can MOPO solve tasks that require generalization to out-of-distribution behaviors? (3) How does each component in MOPO affect performance? 

Question (2) is particularly relevant for scenarios in which we have logged interactions with the environment but want to use those data to optimize a policy for a different reward function.
To study (2) and challenge methods further, we construct two additional continuous control tasks that demand out-of-distribution generalization, as described in Section~\ref{sec:ood}.
To answer question (3), we conduct a complete ablation study to analyze the effect of each module in MOPO in Appendix~\ref{app:ablation}.
For more details on the experimental set-up and hyperparameters, see Appendix~\ref{app:details}.
For more details on the experimental set-up and hyperparameters, see Appendix~\ref{app:details}. The code is available online\footnote{Code is released at \url{https://github.com/tianheyu927/mopo}.}.

We compare against several baselines, including the current state-of-the-art model-free offline RL algorithms.
Bootstrapping error accumulation reduction (BEAR) aims to constrain the policy's actions to lie in the support of the behavioral distribution \cite{kumar2019stabilizing}. This is implemented as a constraint on the average MMD \cite{gretton2007kernel} between $\pi(\cdot \given s)$ and a generative model that approximates $\pib(\cdot \given s)$.
Behavior-regularized actor critic (BRAC) is a family of algorithms that operate by penalizing the value function by some measure of discrepancy (KL divergence or MMD) between $\pi(\cdot \given s)$ and $\pib(\cdot \given s)$ \cite{wu2019behavior}.
BRAC-v uses this penalty both when updating the critic and when updating the actor, while BRAC-p uses this penalty only when updating the actor and does not explicitly penalize the critic.

\subsection{Evaluation on the D4RL benchmark}

\begin{table}[t]
\centering
\small
\begin{tabular}{l|l|r|r|r|r|r|r}
\toprule
\textbf{\!\!\!Dataset type\!\!} & \textbf{Environment} & \textbf{BC} & \textbf{MOPO (ours)} & \textbf{MBPO} & \textbf{SAC} & \textbf{BEAR} & \textbf{BRAC-v}\\ \midrule
\!\!\!random & halfcheetah & 2.1  & \textbf{35.4} $\pm$ 2.5 & 30.7 $\pm$ 3.9 & 30.5 & 25.5 & 28.1 \\
\!\!\!random & hopper & 1.6 & 11.7 $\pm$ 0.4 & 4.5 $\pm$ 6.0 & 11.3 & 9.5 & \textbf{12.0} \\
\!\!\!random & walker2d & 9.8 & \textbf{13.6} $\pm$ 2.6 & 8.6 $\pm$ 8.1 & 4.1 & 6.7 & 0.5 \\
\!\!\!medium & halfcheetah & 36.1 & 42.3 $\pm$ 1.6 & 28.3 $\pm$ 22.7 & -4.3 & 38.6 & \textbf{45.5} \\
\!\!\!medium & hopper & 29.0 & 28.0 $\pm$ 12.4 & 4.9 $\pm$ 3.3 & 0.8 & \textbf{47.6} & 32.3 \\
\!\!\!medium & walker2d & 6.6 & 17.8 $\pm$ 19.3 & 12.7 $\pm$ 7.6 & 0.9 & 33.2 & \textbf{81.3} \\
\!\!\!mixed & halfcheetah & 38.4 & \textbf{53.1} $\pm$ 2.0 & 47.3 $\pm$ 12.6 & -2.4 & 36.2 & 45.9 \\
\!\!\!mixed & hopper & 11.8 & \textbf{67.5} $\pm$ 24.7 & 49.8 $\pm$ 30.4 & 1.9 & 10.8 & 0.9 \\
\!\!\!mixed & walker2d & 11.3 & \textbf{39.0} $\pm$ 9.6 & 22.2 $\pm$ 12.7 & 3.5 & 25.3 & 0.8 \\
\!\!\!med-expert\!\!\! & halfcheetah & 35.8 & \textbf{63.3} $\pm$38.0 & 9.7 $\pm$ 9.5 & 1.8 & 51.7 & 45.3 \\
\!\!\!med-expert\!\!\! & hopper & 111.9 & 23.7 $\pm$ 6.0 & \textbf{56.0} $\pm$ 34.5 & 1.6 & 4.0 & 0.8 \\
\!\!\!med-expert\!\!\! & walker2d & 6.4 & 44.6 $\pm$ 12.9 & 7.6 $\pm$ 3.7 & -0.1 & 26.0 & \textbf{66.6}\\
\bottomrule
\end{tabular}
\vspace{0.1cm}
\caption{\footnotesize Results for D4RL datasets. %
Each number is the normalized score proposed in \cite{fu2020d4rl} of the policy at the last iteration of training, averaged over 6 random seeds, $\pm$ standard deviation. The scores are undiscounted average returns normalized to roughly lie between 0 and 100, where a score of 0 corresponds to
a random policy, and 100 corresponds to an expert.
We include the performance of behavior cloning (\textbf{BC}) from the batch data for comparison.
Numbers for model-free methods taken from \cite{fu2020d4rl}, which does not report standard deviation.
We omit BRAC-p in this table for space because BRAC-v obtains higher performance in 10 of these 12 tasks and is only slightly weaker on the other two.
We bold the highest mean.
}
\label{fig:d4rl}
\normalsize
\vspace{-0.8cm}
\end{table}

To answer question (1), we evaluate our method on a large subset of datasets in the \href{https://sites.google.com/view/d4rl}{D4RL benchmark} \cite{fu2020d4rl} based on the MuJoCo simulator~\citep{todorov2012mujoco}, %
including three environments (halfcheetah, hopper, and walker2d) and four dataset types (random, medium, mixed, medium-expert), yielding a total of 12 problem settings. We also perform empirical evaluations on non-MuJoCo environments in Appendix~\ref{app:hiv}.
The datasets in this benchmark have been generated as follows:
\textbf{random}: roll out a randomly initialized policy for 1M steps.
\textbf{medium}: partially train a policy using SAC, then roll it out for 1M steps.
\textbf{mixed}: train a policy using SAC until a certain (environment-specific) performance threshold is reached, and take the replay buffer as the batch.
\textbf{medium-expert}: combine 1M samples of rollouts from a fully-trained policy with another 1M samples of rollouts from a partially trained policy or a random policy.

Results are given in Table~\ref{fig:d4rl}.
MOPO is the strongest by a significant margin on all the mixed datasets and most of the medium-expert datasets, while also achieving strong performance on all of the random datasets. 
MOPO performs less well on the medium datasets. We hypothesize that the lack of action diversity in the medium datasets make it more difficult to learn a model that generalizes well. Fortunately, this setting is one in which model-free methods can perform well, suggesting that model-based and model-free approaches are able to perform well in complementary settings.

\subsection{Evaluation on tasks requiring out-of-distribution generalization %
}
\label{sec:ood}

To answer question (2), we construct two environments \texttt{halfcheetah-jump} and \texttt{ant-angle} where the agent must solve a task that is different from the purpose of the behavioral policy. 
The trajectories of the batch data in the these datasets are from policies trained for the original dynamics and reward functions \texttt{HalfCheetah} and \texttt{Ant} in OpenAI Gym~\cite{brockman2016openai} which incentivize the cheetach and ant to move forward as fast as possible. Note that for \texttt{HalfCheetah}, we set the maximum velocity to be $3$. Concretely, we train SAC for 1M steps and use the entire training replay buffer as the trajectories for the batch data. Then, we assign these trajectories with new rewards that incentivize the cheetach to jump and the ant to run towards the top right corner with a 30 degree angle. Thus, to achieve good performance for the new reward functions, the policy need to leave the observational distribution, as visualized in Figure~\ref{fig:generalization_visual}. We include the exact forms of the new reward functions in Appendix~\ref{app:details}.
In these environments, learning the correct behaviors requires leaving the support of the data distribution;  optimizing solely within the data manifold will lead to sub-optimal policies.
In Table~\ref{fig:generalize}, we show that MOPO significantly outperforms the state-of-the-art model-free approaches. In particular, model-free offline RL cannot outperform the best trajectory in the batch dataset, whereas MOPO exceeds the batch max by a significant margin.
This validates that MOPO is able to generalize to out-of-distribution behaviors while existing model-free methods are unable to solve those challenges. Note that vanilla MBPO performs much better than SAC in the two environments, consolidating our claim that vanilla model-based methods can attain better results than model-free methods in the offline setting, especially where generalization to out-of-distribution is needed. The visualization in Figure~\ref{fig:generalization_visual} suggests indeed the policy learned MOPO can effectively solve the tasks by reaching to states unseen in the batch data. Furthermore, we test the limit of the generalization abilities of MOPO in these environments and the results are included in Appendix~\ref{app:generalization_limit}.%

\begin{figure}[t]
    \centering
    \includegraphics[width=0.9\linewidth]{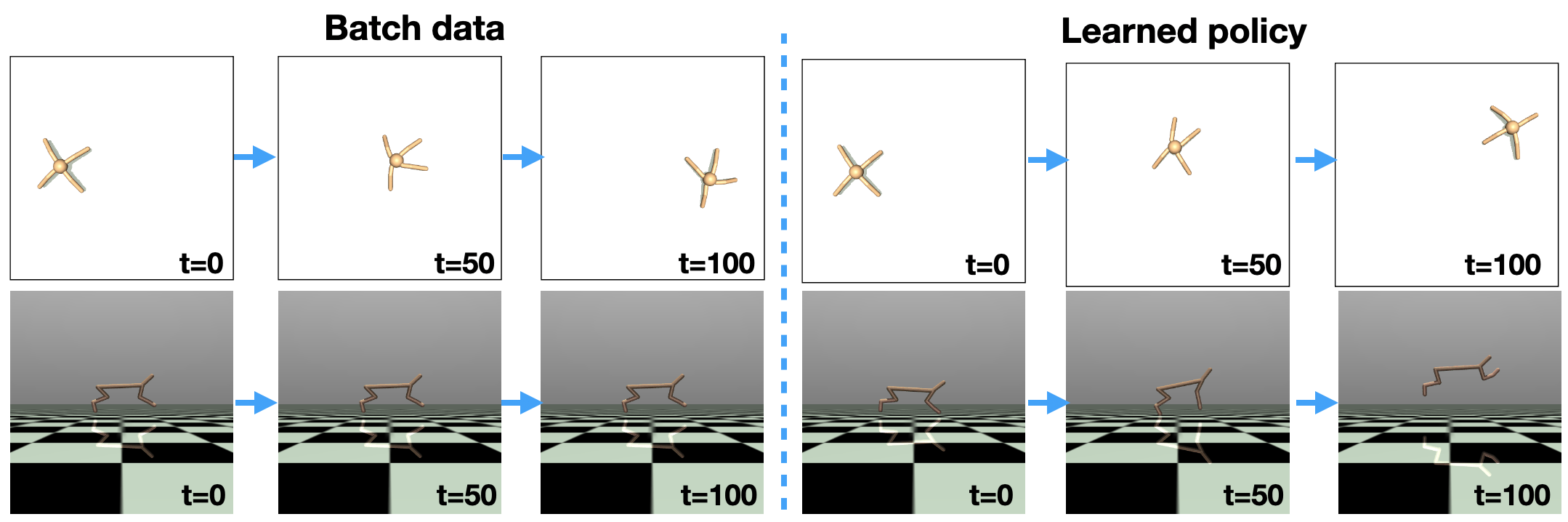}
    \vspace{-0.2cm}
    \caption{\footnotesize
    We visualize the two out-of-distribution generalization environments \texttt{halfcheetah-jump} (\textbf{bottom row}) and \texttt{ant-angle} (\textbf{top row}). We show the training environments that generate the batch data on the left. On the right, we show the test environments where the agents perform behaviors that require the learned policies to leave the data support. In \texttt{halfcheetah-jump}, the agent is asked to run while jumping as high as possible given an training offline dataset of halfcheetah running. In \texttt{ant-angle}, the ant is rewarded for running forward in a 30 degree angle and the corresponding training offline dataset contains data of the ant running forward directly.}
    \label{fig:generalization_visual}
    \vspace{-0.3cm}
\end{figure}

\begin{table}[h]
\centering
\tiny
\begin{tabular}{l|r|r|r|r|r|r|r|r}
\toprule 
\textbf{Environment} & \stackanchor{\textbf{Batch}}{\textbf{Mean}} & \stackanchor{\textbf{Batch}}{\textbf{Max}} & \textbf{MOPO (ours)} & \textbf{MBPO} & \textbf{SAC} & \textbf{BEAR} & \textbf{BRAC-p} & \textbf{BRAC-v}\\ \midrule
halfcheetah-jump & -1022.6 & 1808.6 & \textbf{4016.6}$\mathbf{\pm}$\textbf{144} & 2971.4$\pm$1262 & -3588.2$\pm$1436 & 16.8$\pm$60 & 1069.9$\pm$232 & 871$\pm$41\\
ant-angle & 866.7 & 2311.9 & \textbf{2530.9}$\mathbf{\pm}$\textbf{137} & 13.6$\pm$66 & -966.4$\pm$778 & 1658.2$\pm$16 & 1806.7$\pm$265 & 2333$\pm$139\\
\bottomrule
\end{tabular}
\caption{
\small Average returns \texttt{halfcheetah-jump} and \texttt{ant-angle} that require out-of-distribution policy. The MOPO results are averaged over 6 random seeds, $\pm$ standard deviation, while the results of other methods are averaged over 3 random seeds. We include the mean and max undiscounted return of the episodes in the batch data (under Batch Mean and Batch Max, respectively) for comparison. Note that Batch Mean and Max are significantly lower than on-policy SAC, suggesting that the behaviors stored in the buffers are far from optimal and the agent needs to go beyond the data support in order to achieve better performance. As shown in the results, MOPO outperforms all the baselines by a large margin, indicating that MOPO is effective in generalizing to out-of-distribution states where model-free offline RL methods struggle.}
\vspace{-0.7cm}
\label{fig:generalize}
\normalsize
\end{table}

\section{Conclusion}

In this paper, we studied model-based offline RL algorithms. We started with the observation that, in the offline setting, existing model-based methods significantly outperform vanilla model-free methods, suggesting that model-based methods are more resilient to the overestimation and overfitting issues that plague off-policy model-free RL algorithms. This phenomenon implies that model-based RL has the ability to generalize to states outside of the data support and such generalization is conducive for offline RL. 
However, online and offline algorithms must act differently when handling out-of-distribution states. Model error on out-of-distribution states that often drives exploration and corrective feedback in the online setting~\cite{kumar2020discor} can be detrimental when interaction is not allowed. Using theoretical principles, we develop an algorithm, model-based offline policy optimization (MOPO), which maximizes the policy on a MDP that penalizes states with high model uncertainty. MOPO trades off the risk of making mistakes and the benefit of diverse exploration from escaping the behavioral distribution. In our experiments, MOPO outperforms state-of-the-art offline RL methods in both standard benchmarks~\cite{fu2020d4rl} and out-of-distribution generalization environments.

Our work opens up a number of questions and directions for future work. First, an interesting avenue for future research to incorporate the policy regularization ideas of BEAR and BRAC into the reward penalty framework to improve the performance of MOPO on narrow data distributions (such as the ``medium'' datasets in D4RL). 
Second, it's an interesting theoretical question to understand why model-based methods appear to be much better suited to the batch setting than model-free methods. Multiple potential factors include a greater supervision from the states (instead of only the reward), more stable and less noisy supervised gradient updates, or ease of uncertainty estimation. Our work suggests that uncertainty estimation plays an important role, particularly in settings that demand generalization. However, uncertainty estimation does not explain the entire difference nor does it explain why model-free methods cannot also enjoy the benefits of uncertainty estimation. For those domains where learning a model may be very difficult due to complex dynamics, developing better model-free offline RL methods may be desirable or imperative. Hence, it is crucial to conduct future research on investigating how to bring model-free offline RL methods up to the level of the performance of model-based methods, which would require further understanding where the generalization benefits come from.

\section*{Broader Impact}
MOPO achieves significant strides in offline reinforcement learning, a problem setting that is particularly scalable to real-world settings.
Offline reinforcement learning has a number of potential application domains, including autonomous driving, healthcare, robotics, and is notably amenable to safety-critical settings where online data collection is costly.
For example, in autonomous driving, online interaction with the environment runs the risk of crashing and hurting people; offline RL methods can significantly reduce that risk by learning from a pre-recorded driving dataset collected by a safe behavioral policy.
Moreover, our work opens up the possibility of learning policies offline for new tasks for which we do not already have expert data.

However, there are still risks associated with applying learned policies to high-risk domains.
We have shown the benefits of explicitly accounting for error, but without reliable out-of-distribution uncertainty estimation techniques, there is a possibility that the policy will behave unpredictably when given a scenario it has not encountered.
There is also the challenge of reward design: although the reward function will typically be under the engineer's control, it can be difficult to specify a reward function that elicits the desired behavior and is aligned with human objectives.
Additionally, parametric models are known to be susceptible to adversarial attacks, and bad actors can potentially exploit this vulnerability.
Advances in uncertainty quantification, human-computer interaction, and robustness will improve our ability to apply learning-based methods in safety-critical domains.

Supposing we succeed at producing safe and reliable policies, there is still possibility of negative societal impact.
An increased ability to automate decision-making processes may reduce companies' demand for employees in certain industries (e.g. manufacturing and logistics), thereby affecting job availability.
However, historically, advances in technology have also created new jobs that did not previously exist (e.g. software engineering), and it is unclear if the net impact on jobs will be positive or negative.

Despite the aforementioned risks and challenges, we believe that offline RL is a promising setting with enormous potential for automating and improving sequential decision-making in highly impactful domains.
Currently, much additional work is needed to make offline RL sufficiently robust to be applied in safety-critical settings.
We encourage the research community to pursue further study in uncertainty estimation, particularly considering the complications that arise in sequential decision problems.

\begin{ack}

We thank Michael Janner for help with MBPO and Aviral Kumar for setting up BEAR and D4RL. TY is partially supported by Intel Corporation. CF is a CIFAR Fellow in the Learning in Machines and Brains program. TM and GT are also partially supported by Lam Research, Google Faculty Award, SDSI, and SAIL.
\end{ack}

\bibliography{reference}

\begin{thebibliography}{76}
\providecommand{\natexlab}[1]{#1}
\providecommand{\url}[1]{\texttt{#1}}
\expandafter\ifx\csname urlstyle\endcsname\relax
  \providecommand{\doi}[1]{doi: #1}\else
  \providecommand{\doi}{doi: \begingroup \urlstyle{rm}\Url}\fi

\bibitem[Agarwal et~al.(2019)Agarwal, Schuurmans, and
  Norouzi]{agarwal2019striving}
Rishabh Agarwal, Dale Schuurmans, and Mohammad Norouzi.
\newblock Striving for simplicity in off-policy deep reinforcement learning.
\newblock \emph{arXiv preprint arXiv:1907.04543}, 2019.

\bibitem[Banaszuk et~al.(2011)Banaszuk, Fonoberov, Frewen, Kobilarov, Mathew,
  Mezic, Pinto, Sahai, Sane, Speranzon, et~al.]{banaszuk2011scalable}
Andrzej Banaszuk, Vladimir~A Fonoberov, Thomas~A Frewen, Marin Kobilarov,
  George Mathew, Igor Mezic, Alessandro Pinto, Tuhin Sahai, Harshad Sane,
  Alberto Speranzon, et~al.
\newblock Scalable approach to uncertainty quantification and robust design of
  interconnected dynamical systems.
\newblock \emph{Annual Reviews in Control}, 35\penalty0 (1):\penalty0 77--98,
  2011.

\bibitem[Barto et~al.(1983)Barto, Sutton, and Anderson]{barto1983neuronlike}
Andrew~G Barto, Richard~S Sutton, and Charles~W Anderson.
\newblock Neuronlike adaptive elements that can solve difficult learning
  control problems.
\newblock \emph{IEEE transactions on systems, man, and cybernetics}, \penalty0
  (5):\penalty0 834--846, 1983.

\bibitem[Bengio et~al.(2020)Bengio, Pineau, and Precup]{bengio2020interference}
Emmanuel Bengio, Joelle Pineau, and Doina Precup.
\newblock Interference and generalization in temporal difference learning.
\newblock \emph{arXiv preprint arXiv:2003.06350}, 2020.

\bibitem[Bickel and Freedman(1981)]{bickel1981some}
Peter~J Bickel and David~A Freedman.
\newblock Some asymptotic theory for the bootstrap.
\newblock \emph{The annals of statistics}, pages 1196--1217, 1981.

\bibitem[Brockman et~al.(2016)Brockman, Cheung, Pettersson, Schneider,
  Schulman, Tang, and Zaremba]{brockman2016openai}
Greg Brockman, Vicki Cheung, Ludwig Pettersson, Jonas Schneider, John Schulman,
  Jie Tang, and Wojciech Zaremba.
\newblock Openai gym.
\newblock \emph{arXiv preprint arXiv:1606.01540}, 2016.

\bibitem[Chua et~al.(2018)Chua, Calandra, McAllister, and Levine]{chua2018deep}
Kurtland Chua, Roberto Calandra, Rowan McAllister, and Sergey Levine.
\newblock Deep reinforcement learning in a handful of trials using
  probabilistic dynamics models.
\newblock In \emph{Advances in Neural Information Processing Systems}, pages
  4754--4765, 2018.

\bibitem[Clavera et~al.(2018)Clavera, Rothfuss, Schulman, Fujita, Asfour, and
  Abbeel]{clavera2018model}
Ignasi Clavera, Jonas Rothfuss, John Schulman, Yasuhiro Fujita, Tamim Asfour,
  and Pieter Abbeel.
\newblock Model-based reinforcement learning via meta-policy optimization.
\newblock \emph{arXiv preprint arXiv:1809.05214}, 2018.

\bibitem[Cobbe et~al.(2018)Cobbe, Klimov, Hesse, Kim, and
  Schulman]{cobbe2018quantifying}
Karl Cobbe, Oleg Klimov, Chris Hesse, Taehoon Kim, and John Schulman.
\newblock Quantifying generalization in reinforcement learning.
\newblock \emph{arXiv preprint arXiv:1812.02341}, 2018.

\bibitem[Dasari et~al.(2019)Dasari, Ebert, Tian, Nair, Bucher, Schmeckpeper,
  Singh, Levine, and Finn]{dasari2019robonet}
Sudeep Dasari, Frederik Ebert, Stephen Tian, Suraj Nair, Bernadette Bucher,
  Karl Schmeckpeper, Siddharth Singh, Sergey Levine, and Chelsea Finn.
\newblock Robonet: Large-scale multi-robot learning.
\newblock \emph{arXiv preprint arXiv:1910.11215}, 2019.

\bibitem[Degris et~al.(2012)Degris, White, and Sutton]{degris2012off}
Thomas Degris, Martha White, and Richard~S Sutton.
\newblock Off-policy actor-critic.
\newblock \emph{arXiv preprint arXiv:1205.4839}, 2012.

\bibitem[Deisenroth and Rasmussen(2011)]{deisenroth2011pilco}
Marc Deisenroth and Carl~E Rasmussen.
\newblock Pilco: A model-based and data-efficient approach to policy search.
\newblock In \emph{Proceedings of the 28th International Conference on machine
  learning (ICML-11)}, pages 465--472, 2011.

\bibitem[Deng et~al.(2009)Deng, Dong, Socher, Li, Li, and
  Fei-Fei]{deng2009imagenet}
Jia Deng, Wei Dong, Richard Socher, Li-Jia Li, Kai Li, and Li~Fei-Fei.
\newblock Imagenet: A large-scale hierarchical image database.
\newblock In \emph{2009 IEEE conference on computer vision and pattern
  recognition}, pages 248--255. Ieee, 2009.

\bibitem[Depeweg et~al.(2016)Depeweg, Hern{\'a}ndez-Lobato, Doshi-Velez, and
  Udluft]{depeweg2016learning}
Stefan Depeweg, Jos{\'e}~Miguel Hern{\'a}ndez-Lobato, Finale Doshi-Velez, and
  Steffen Udluft.
\newblock Learning and policy search in stochastic dynamical systems with
  bayesian neural networks.
\newblock \emph{arXiv preprint arXiv:1605.07127}, 2016.

\bibitem[Draeger et~al.(1995)Draeger, Engell, and Ranke]{draeger1995model}
Andreas Draeger, Sebastian Engell, and Horst Ranke.
\newblock Model predictive control using neural networks.
\newblock \emph{IEEE Control Systems Magazine}, 15\penalty0 (5):\penalty0
  61--66, 1995.

\bibitem[Ebert et~al.(2018)Ebert, Finn, Dasari, Xie, Lee, and
  Levine]{ebert2018visual}
Frederik Ebert, Chelsea Finn, Sudeep Dasari, Annie Xie, Alex Lee, and Sergey
  Levine.
\newblock Visual foresight: Model-based deep reinforcement learning for
  vision-based robotic control.
\newblock \emph{arXiv preprint arXiv:1812.00568}, 2018.

\bibitem[Finn and Levine(2017)]{finn2017deep}
Chelsea Finn and Sergey Levine.
\newblock Deep visual foresight for planning robot motion.
\newblock In \emph{2017 IEEE International Conference on Robotics and
  Automation (ICRA)}, pages 2786--2793. IEEE, 2017.

\bibitem[Fu et~al.(2020)Fu, Kumar, Nachum, Tucker, and Levine]{fu2020d4rl}
Justin Fu, Aviral Kumar, Ofir Nachum, George Tucker, and Sergey Levine.
\newblock D4rl: Datasets for deep data-driven reinforcement learning, 2020.

\bibitem[Fujimoto et~al.(2018{\natexlab{a}})Fujimoto, Meger, and
  Precup]{fujimoto2018off}
Scott Fujimoto, David Meger, and Doina Precup.
\newblock Off-policy deep reinforcement learning without exploration.
\newblock \emph{arXiv preprint arXiv:1812.02900}, 2018{\natexlab{a}}.

\bibitem[Fujimoto et~al.(2018{\natexlab{b}})Fujimoto, Van~Hoof, and
  Meger]{fujimoto2018addressing}
Scott Fujimoto, Herke Van~Hoof, and David Meger.
\newblock Addressing function approximation error in actor-critic methods.
\newblock \emph{arXiv preprint arXiv:1802.09477}, 2018{\natexlab{b}}.

\bibitem[Gal et~al.(2016)Gal, McAllister, and Rasmussen]{gal2016improving}
Yarin Gal, Rowan McAllister, and Carl~Edward Rasmussen.
\newblock Improving pilco with bayesian neural network dynamics models.
\newblock In \emph{Data-Efficient Machine Learning workshop, ICML}, volume~4,
  page~34, 2016.

\bibitem[Gottesman et~al.(2019)Gottesman, Johansson, Komorowski, Faisal,
  Sontag, Doshi-Velez, and Celi]{gottesman2019guidelines}
Omer Gottesman, Fredrik Johansson, Matthieu Komorowski, Aldo Faisal, David
  Sontag, Finale Doshi-Velez, and Leo~Anthony Celi.
\newblock Guidelines for reinforcement learning in healthcare.
\newblock \emph{Nat Med}, 25\penalty0 (1):\penalty0 16--18, 2019.

\bibitem[Gretton et~al.(2007)Gretton, Borgwardt, Rasch, Scholk\"{o}pf, and
  Smola]{gretton2007kernel}
Arthur Gretton, Karsten~M. Borgwardt, Malte Rasch, Bernhard Scholk\"{o}pf, and
  Alexander~J. Smola.
\newblock A kernel approach to comparing distributions.
\newblock In \emph{Proceedings of the 22nd National Conference on Artificial
  Intelligence - Volume 2}, AAAI’07, page 1637–1641. AAAI Press, 2007.
\newblock ISBN 9781577353232.

\bibitem[Gu et~al.(2016)Gu, Lillicrap, Ghahramani, Turner, and Levine]{gu2016q}
Shixiang Gu, Timothy Lillicrap, Zoubin Ghahramani, Richard~E Turner, and Sergey
  Levine.
\newblock Q-prop: Sample-efficient policy gradient with an off-policy critic.
\newblock \emph{arXiv preprint arXiv:1611.02247}, 2016.

\bibitem[Gu et~al.(2017)Gu, Lillicrap, Turner, Ghahramani, Sch{\"o}lkopf, and
  Levine]{gu2017interpolated}
Shixiang~Shane Gu, Timothy Lillicrap, Richard~E Turner, Zoubin Ghahramani,
  Bernhard Sch{\"o}lkopf, and Sergey Levine.
\newblock Interpolated policy gradient: Merging on-policy and off-policy
  gradient estimation for deep reinforcement learning.
\newblock In \emph{Advances in neural information processing systems}, pages
  3846--3855, 2017.

\bibitem[Ha and Schmidhuber(2018)]{ha2018world}
David Ha and J{\"u}rgen Schmidhuber.
\newblock World models.
\newblock \emph{arXiv preprint arXiv:1803.10122}, 2018.

\bibitem[Haarnoja et~al.(2018)Haarnoja, Zhou, Abbeel, and
  Levine]{haarnoja2018soft}
Tuomas Haarnoja, Aurick Zhou, Pieter Abbeel, and Sergey Levine.
\newblock Soft actor-critic: Off-policy maximum entropy deep reinforcement
  learning with a stochastic actor.
\newblock \emph{arXiv preprint arXiv:1801.01290}, 2018.

\bibitem[Holland et~al.(2018)Holland, Talvitie, and Bowling]{holland2018effect}
G~Zacharias Holland, Erin~J Talvitie, and Michael Bowling.
\newblock The effect of planning shape on dyna-style planning in
  high-dimensional state spaces.
\newblock \emph{arXiv preprint arXiv:1806.01825}, 2018.

\bibitem[Janner et~al.(2019)Janner, Fu, Zhang, and Levine]{janner2019trust}
Michael Janner, Justin Fu, Marvin Zhang, and Sergey Levine.
\newblock When to trust your model: Model-based policy optimization.
\newblock In \emph{Advances in Neural Information Processing Systems}, pages
  12498--12509, 2019.

\bibitem[Jaques et~al.(2019)Jaques, Ghandeharioun, Shen, Ferguson, Lapedriza,
  Jones, Gu, and Picard]{jaques2019way}
Natasha Jaques, Asma Ghandeharioun, Judy~Hanwen Shen, Craig Ferguson, Agata
  Lapedriza, Noah Jones, Shixiang Gu, and Rosalind Picard.
\newblock Way off-policy batch deep reinforcement learning of implicit human
  preferences in dialog.
\newblock \emph{arXiv preprint arXiv:1907.00456}, 2019.

\bibitem[Jiang and Li(2015)]{jiang2015doubly}
Nan Jiang and Lihong Li.
\newblock Doubly robust off-policy value evaluation for reinforcement learning.
\newblock \emph{arXiv preprint arXiv:1511.03722}, 2015.

\bibitem[Kaiser et~al.(2019)Kaiser, Babaeizadeh, Milos, Osinski, Campbell,
  Czechowski, Erhan, Finn, Kozakowski, Levine, Mohiuddin, Sepassi, Tucker, and
  Michalewski]{kaiser2019modelbased}
Lukasz Kaiser, Mohammad Babaeizadeh, Piotr Milos, Blazej Osinski, Roy~H
  Campbell, Konrad Czechowski, Dumitru Erhan, Chelsea Finn, Piotr Kozakowski,
  Sergey Levine, Afroz Mohiuddin, Ryan Sepassi, George Tucker, and Henryk
  Michalewski.
\newblock Model-based reinforcement learning for atari, 2019.

\bibitem[Kidambi et~al.(2020)Kidambi, Rajeswaran, Netrapalli, and
  Joachims]{kidambi2020morel}
Rahul Kidambi, Aravind Rajeswaran, Praneeth Netrapalli, and Thorsten Joachims.
\newblock Morel: Model-based offline reinforcement learning.
\newblock \emph{arXiv preprint arXiv:2005.05951}, 2020.

\bibitem[Kim et~al.(2013)Kim, Shen, Nagy, and Braatz]{kim2013wiener}
Kwang-Ki~K Kim, Dongying~Erin Shen, Zoltan~K Nagy, and Richard~D Braatz.
\newblock Wiener's polynomial chaos for the analysis and control of nonlinear
  dynamical systems with probabilistic uncertainties [historical perspectives].
\newblock \emph{IEEE Control Systems Magazine}, 33\penalty0 (5):\penalty0
  58--67, 2013.

\bibitem[Kuleshov et~al.(2018)Kuleshov, Fenner, and
  Ermon]{kuleshov2018accurate}
Volodymyr Kuleshov, Nathan Fenner, and Stefano Ermon.
\newblock Accurate uncertainties for deep learning using calibrated regression.
\newblock \emph{arXiv preprint arXiv:1807.00263}, 2018.

\bibitem[Kumar et~al.(2019)Kumar, Fu, Soh, Tucker, and
  Levine]{kumar2019stabilizing}
Aviral Kumar, Justin Fu, Matthew Soh, George Tucker, and Sergey Levine.
\newblock Stabilizing off-policy q-learning via bootstrapping error reduction.
\newblock In \emph{Advances in Neural Information Processing Systems}, pages
  11761--11771, 2019.

\bibitem[Kumar et~al.(2020)Kumar, Gupta, and Levine]{kumar2020discor}
Aviral Kumar, Abhishek Gupta, and Sergey Levine.
\newblock Discor: Corrective feedback in reinforcement learning via
  distribution correction.
\newblock \emph{arXiv preprint arXiv:2003.07305}, 2020.

\bibitem[Kumar et~al.(2016)Kumar, Todorov, and Levine]{kumar2016optimal}
Vikash Kumar, Emanuel Todorov, and Sergey Levine.
\newblock Optimal control with learned local models: Application to dexterous
  manipulation.
\newblock In \emph{2016 IEEE International Conference on Robotics and
  Automation (ICRA)}, pages 378--383. IEEE, 2016.

\bibitem[Kurutach et~al.(2018)Kurutach, Clavera, Duan, Tamar, and
  Abbeel]{kurutach2018model}
Thanard Kurutach, Ignasi Clavera, Yan Duan, Aviv Tamar, and Pieter Abbeel.
\newblock Model-ensemble trust-region policy optimization.
\newblock \emph{arXiv preprint arXiv:1802.10592}, 2018.

\bibitem[Lakshminarayanan et~al.(2017)Lakshminarayanan, Pritzel, and
  Blundell]{lakshminarayanan2017simple}
Balaji Lakshminarayanan, Alexander Pritzel, and Charles Blundell.
\newblock Simple and scalable predictive uncertainty estimation using deep
  ensembles.
\newblock In \emph{Advances in neural information processing systems}, pages
  6402--6413, 2017.

\bibitem[Lange et~al.(2012)Lange, Gabel, and Riedmiller]{lange2012batch}
Sascha Lange, Thomas Gabel, and Martin Riedmiller.
\newblock Batch reinforcement learning.
\newblock In \emph{Reinforcement learning}, pages 45--73. Springer, 2012.

\bibitem[Levine and Koltun(2013)]{levine2013guided}
Sergey Levine and Vladlen Koltun.
\newblock Guided policy search.
\newblock In \emph{International Conference on Machine Learning}, pages 1--9,
  2013.

\bibitem[Lillicrap et~al.(2015)Lillicrap, Hunt, Pritzel, Heess, Erez, Tassa,
  Silver, and Wierstra]{lillicrap2015continuous}
Timothy~P Lillicrap, Jonathan~J Hunt, Alexander Pritzel, Nicolas Heess, Tom
  Erez, Yuval Tassa, David Silver, and Daan Wierstra.
\newblock Continuous control with deep reinforcement learning.
\newblock \emph{arXiv preprint arXiv:1509.02971}, 2015.

\bibitem[Luo et~al.(2018)Luo, Xu, Li, Tian, Darrell, and
  Ma]{luo2018algorithmic}
Yuping Luo, Huazhe Xu, Yuanzhi Li, Yuandong Tian, Trevor Darrell, and Tengyu
  Ma.
\newblock Algorithmic framework for model-based deep reinforcement learning
  with theoretical guarantees.
\newblock \emph{arXiv preprint arXiv:1807.03858}, 2018.

\bibitem[Miyato et~al.(2018)Miyato, Kataoka, Koyama, and
  Yoshida]{miyato2018spectral}
Takeru Miyato, Toshiki Kataoka, Masanori Koyama, and Yuichi Yoshida.
\newblock Spectral normalization for generative adversarial networks.
\newblock \emph{arXiv preprint arXiv:1802.05957}, 2018.

\bibitem[Mnih et~al.(2016)Mnih, Badia, Mirza, Graves, Lillicrap, Harley,
  Silver, and Kavukcuoglu]{mnih2016asynchronous}
Volodymyr Mnih, Adria~Puigdomenech Badia, Mehdi Mirza, Alex Graves, Timothy
  Lillicrap, Tim Harley, David Silver, and Koray Kavukcuoglu.
\newblock Asynchronous methods for deep reinforcement learning.
\newblock In \emph{International conference on machine learning}, pages
  1928--1937, 2016.

\bibitem[M{\"u}ller(1997)]{muller1997integral}
Alfred M{\"u}ller.
\newblock Integral probability metrics and their generating classes of
  functions.
\newblock \emph{Advances in Applied Probability}, 29\penalty0 (2):\penalty0
  429--443, 1997.

\bibitem[Munos et~al.(2016)Munos, Stepleton, Harutyunyan, and
  Bellemare]{munos2016safe}
R{\'e}mi Munos, Tom Stepleton, Anna Harutyunyan, and Marc Bellemare.
\newblock Safe and efficient off-policy reinforcement learning.
\newblock In \emph{Advances in Neural Information Processing Systems}, pages
  1054--1062, 2016.

\bibitem[Nachum et~al.(2019)Nachum, Dai, Kostrikov, Chow, Li, and
  Schuurmans]{nachum2019algaedice}
Ofir Nachum, Bo~Dai, Ilya Kostrikov, Yinlam Chow, Lihong Li, and Dale
  Schuurmans.
\newblock Algaedice: Policy gradient from arbitrary experience.
\newblock \emph{arXiv preprint arXiv:1912.02074}, 2019.

\bibitem[Nagabandi et~al.(2019)Nagabandi, Konoglie, Levine, and
  Kumar]{nagabandi2019deep}
Anusha Nagabandi, Kurt Konoglie, Sergey Levine, and Vikash Kumar.
\newblock Deep dynamics models for learning dexterous manipulation.
\newblock \emph{arXiv preprint arXiv:1909.11652}, 2019.

\bibitem[Oh et~al.(2017)Oh, Singh, and Lee]{oh2017value}
Junhyuk Oh, Satinder Singh, and Honglak Lee.
\newblock Value prediction network.
\newblock In \emph{Advances in Neural Information Processing Systems}, pages
  6118--6128, 2017.

\bibitem[Peng et~al.(2019)Peng, Kumar, Zhang, and Levine]{peng2019advantage}
Xue~Bin Peng, Aviral Kumar, Grace Zhang, and Sergey Levine.
\newblock Advantage-weighted regression: Simple and scalable off-policy
  reinforcement learning.
\newblock \emph{arXiv preprint arXiv:1910.00177}, 2019.

\bibitem[Precup et~al.(2001)Precup, Sutton, and Dasgupta]{precup2001off}
Doina Precup, Richard~S Sutton, and Sanjoy Dasgupta.
\newblock Off-policy temporal-difference learning with function approximation.
\newblock In \emph{ICML}, pages 417--424, 2001.

\bibitem[Racani{\`e}re et~al.(2017)Racani{\`e}re, Weber, Reichert, Buesing,
  Guez, Rezende, Badia, Vinyals, Heess, Li, et~al.]{racaniere2017imagination}
S{\'e}bastien Racani{\`e}re, Th{\'e}ophane Weber, David Reichert, Lars Buesing,
  Arthur Guez, Danilo~Jimenez Rezende, Adria~Puigdomenech Badia, Oriol Vinyals,
  Nicolas Heess, Yujia Li, et~al.
\newblock Imagination-augmented agents for deep reinforcement learning.
\newblock In \emph{Advances in neural information processing systems}, pages
  5690--5701, 2017.

\bibitem[Rajpurkar et~al.(2016)Rajpurkar, Zhang, Lopyrev, and
  Liang]{rajpurkar2016squad}
Pranav Rajpurkar, Jian Zhang, Konstantin Lopyrev, and Percy Liang.
\newblock Squad: 100,000+ questions for machine comprehension of text.
\newblock \emph{arXiv preprint arXiv:1606.05250}, 2016.

\bibitem[Schulman et~al.(2015)Schulman, Levine, Abbeel, Jordan, and
  Moritz]{schulman2015trust}
John Schulman, Sergey Levine, Pieter Abbeel, Michael Jordan, and Philipp
  Moritz.
\newblock Trust region policy optimization.
\newblock In \emph{International conference on machine learning}, pages
  1889--1897, 2015.

\bibitem[Schulman et~al.(2017)Schulman, Wolski, Dhariwal, Radford, and
  Klimov]{schulman2017proximal}
John Schulman, Filip Wolski, Prafulla Dhariwal, Alec Radford, and Oleg Klimov.
\newblock Proximal policy optimization algorithms.
\newblock \emph{arXiv preprint arXiv:1707.06347}, 2017.

\bibitem[Siegel et~al.(2020)Siegel, Springenberg, Berkenkamp, Abdolmaleki,
  Neunert, Lampe, Hafner, and Riedmiller]{siegel2020keep}
Noah~Y Siegel, Jost~Tobias Springenberg, Felix Berkenkamp, Abbas Abdolmaleki,
  Michael Neunert, Thomas Lampe, Roland Hafner, and Martin Riedmiller.
\newblock Keep doing what worked: Behavioral modelling priors for offline
  reinforcement learning.
\newblock \emph{arXiv preprint arXiv:2002.08396}, 2020.

\bibitem[Silver et~al.(2017)Silver, van Hasselt, Hessel, Schaul, Guez, Harley,
  Dulac-Arnold, Reichert, Rabinowitz, Barreto, et~al.]{silver2017predictron}
David Silver, Hado van Hasselt, Matteo Hessel, Tom Schaul, Arthur Guez, Tim
  Harley, Gabriel Dulac-Arnold, David Reichert, Neil Rabinowitz, Andre Barreto,
  et~al.
\newblock The predictron: End-to-end learning and planning.
\newblock In \emph{Proceedings of the 34th International Conference on Machine
  Learning-Volume 70}, pages 3191--3199. JMLR. org, 2017.

\bibitem[Snoek et~al.(2019)Snoek, Ovadia, Fertig, Lakshminarayanan, Nowozin,
  Sculley, Dillon, Ren, and Nado]{snoek2019can}
Jasper Snoek, Yaniv Ovadia, Emily Fertig, Balaji Lakshminarayanan, Sebastian
  Nowozin, D~Sculley, Joshua Dillon, Jie Ren, and Zachary Nado.
\newblock Can you trust your model's uncertainty? evaluating predictive
  uncertainty under dataset shift.
\newblock In \emph{Advances in Neural Information Processing Systems}, pages
  13969--13980, 2019.

\bibitem[Sriperumbudur et~al.(2009)Sriperumbudur, Fukumizu, Gretton,
  Sch{\"o}lkopf, and Lanckriet]{sriperumbudur2009integral}
Bharath~K Sriperumbudur, Kenji Fukumizu, Arthur Gretton, Bernhard
  Sch{\"o}lkopf, and Gert~RG Lanckriet.
\newblock On integral probability metrics,$\backslash$phi-divergences and
  binary classification.
\newblock \emph{arXiv preprint arXiv:0901.2698}, 2009.

\bibitem[Stengel(1994)]{stengel1994optimal}
Robert~F Stengel.
\newblock \emph{Optimal control and estimation}.
\newblock Courier Corporation, 1994.

\bibitem[Strehl and Littman(2008)]{strehl2008analysis}
Alexander~L Strehl and Michael~L Littman.
\newblock An analysis of model-based interval estimation for markov decision
  processes.
\newblock \emph{Journal of Computer and System Sciences}, 74\penalty0
  (8):\penalty0 1309--1331, 2008.

\bibitem[Sutton(1991)]{sutton1991dyna}
Richard~S Sutton.
\newblock Dyna, an integrated architecture for learning, planning, and
  reacting.
\newblock \emph{ACM Sigart Bulletin}, 2\penalty0 (4):\penalty0 160--163, 1991.

\bibitem[Sutton and Barto(1998)]{sutton1998reinforcement}
Richard~S Sutton and Andrew~G Barto.
\newblock Reinforcement learning, 1998.

\bibitem[Sutton et~al.(2012)Sutton, Szepesv{\'a}ri, Geramifard, and
  Bowling]{sutton2012dyna}
Richard~S Sutton, Csaba Szepesv{\'a}ri, Alborz Geramifard, and Michael~P
  Bowling.
\newblock Dyna-style planning with linear function approximation and
  prioritized sweeping.
\newblock \emph{arXiv preprint arXiv:1206.3285}, 2012.

\bibitem[Tamar et~al.(2016)Tamar, Wu, Thomas, Levine, and
  Abbeel]{tamar2016value}
Aviv Tamar, Yi~Wu, Garrett Thomas, Sergey Levine, and Pieter Abbeel.
\newblock Value iteration networks.
\newblock In \emph{Advances in Neural Information Processing Systems}, pages
  2154--2162, 2016.

\bibitem[Thomas(2015)]{thomas2015safe}
Philip~S Thomas.
\newblock \emph{Safe reinforcement learning}.
\newblock PhD thesis, University of Massachusetts Libraries, 2015.

\bibitem[Todorov et~al.(2012)Todorov, Erez, and Tassa]{todorov2012mujoco}
Emanuel Todorov, Tom Erez, and Yuval Tassa.
\newblock Mujoco: A physics engine for model-based control.
\newblock In \emph{2012 IEEE/RSJ International Conference on Intelligent Robots
  and Systems}, pages 5026--5033. IEEE, 2012.

\bibitem[Wang and Ba(2019)]{wang2019exploring}
Tingwu Wang and Jimmy Ba.
\newblock Exploring model-based planning with policy networks.
\newblock \emph{arXiv preprint arXiv:1906.08649}, 2019.

\bibitem[Wang et~al.(2019)Wang, Bao, Clavera, Hoang, Wen, Langlois, Zhang,
  Zhang, Abbeel, and Ba]{wang2019benchmarking}
Tingwu Wang, Xuchan Bao, Ignasi Clavera, Jerrick Hoang, Yeming Wen, Eric
  Langlois, Shunshi Zhang, Guodong Zhang, Pieter Abbeel, and Jimmy Ba.
\newblock Benchmarking model-based reinforcement learning.
\newblock \emph{arXiv preprint arXiv:1907.02057}, 2019.

\bibitem[Wu et~al.(2019)Wu, Tucker, and Nachum]{wu2019behavior}
Yifan Wu, George Tucker, and Ofir Nachum.
\newblock Behavior regularized offline reinforcement learning.
\newblock \emph{arXiv preprint arXiv:1911.11361}, 2019.

\bibitem[Yao et~al.(2009)Yao, Bhatnagar, Diao, Sutton, and
  Szepesv{\'a}ri]{yao2009multi}
Hengshuai Yao, Shalabh Bhatnagar, Dongcui Diao, Richard~S Sutton, and Csaba
  Szepesv{\'a}ri.
\newblock Multi-step dyna planning for policy evaluation and control.
\newblock In \emph{Advances in neural information processing systems}, pages
  2187--2195, 2009.

\bibitem[Yu et~al.(2018)Yu, Xian, Chen, Liu, Liao, Madhavan, and
  Darrell]{yu2018bdd100k}
Fisher Yu, Wenqi Xian, Yingying Chen, Fangchen Liu, Mike Liao, Vashisht
  Madhavan, and Trevor Darrell.
\newblock Bdd100k: A diverse driving video database with scalable annotation
  tooling.
\newblock \emph{arXiv preprint arXiv:1805.04687}, 2018.

\bibitem[Zanette and Brunskill()]{zanettetighter}
Andrea Zanette and Emma Brunskill.
\newblock Tighter problem-dependent regret bounds in reinforcement learning
  without domain knowledge using value function bounds.

\bibitem[Zhang et~al.(2018)Zhang, Ballas, and Pineau]{zhang2018dissection}
Amy Zhang, Nicolas Ballas, and Joelle Pineau.
\newblock A dissection of overfitting and generalization in continuous
  reinforcement learning.
\newblock \emph{arXiv preprint arXiv:1806.07937}, 2018.

\end{thebibliography}
\bibliographystyle{plainnat}

\newpage
{\Large \bf Appendix}
\appendix

\section{Reminders about integral probability metrics}
\label{app:ipms}
Let $(\mathcal{X}, \Sigma)$ be a measurable space.
The integral probability metric associated with a class $\mathcal{F}$ of (measurable) real-valued functions on $\mathcal{X}$ is defined as
\[d_{\mathcal{F}}(P,Q) = \sup_{f \in \mathcal{F}} \left|\int_{\mathcal{X}} f \dd{P} - \int_{\mathcal{X}} f \dd{Q}\right| = \sup_{f \in \mathcal{F}} \left|\Esub{X \sim P}[f(X)] - \Esub{Y \sim Q}[f(Y)]\right|\]
where $P$ and $Q$ are probability measures on $\mathcal{X}$.
We note the following special cases:
\begin{enumerate}
\item[(i)] If $\mathcal{F} = \{f : \|f\|_\infty \le 1\}$, then $d_{\mathcal{F}}$ is the \textit{total variation distance}
\[d_{\mathcal{F}}(P,Q) = D_{\textsc{tv}}(P,Q) := \sup_{A \in \Sigma} |P(A)-Q(A)|\]

\item[(ii)] If $\mathcal{F}$ is the set of 1-Lipschitz function w.r.t. to some cost function (metric) $c$ on $\mathcal{X}$, then $d_\mathcal{F}$ is the \textit{1-Wasserstein distance} w.r.t. the same metric:
\[d_{\mathcal{F}}(P,Q) = W_1(P,Q) := \inf_{\gamma \in \Gamma(P,Q)} \int_{\mathcal{X}^2} c(x,y) \dd{\gamma(x,y)}\]
where $\Gamma(P,Q)$ denotes the set of all \textit{couplings} of $P$ and $Q$, i.e. joint distributions on $\mathcal{X}^2$ which have marginals $P$ and $Q$.

\item[(iii)] If $\mathcal{F} = \{f : \|f\|_{\mathcal{H}} \le 1\}$ where $\mathcal{H}$ is a reproducing kernel Hilbert space with kernel $k$, then $d_{\mathcal{F}}$ is the \textit{maximum mean discrepancy}:
\[d_{\mathcal{F}}(P,Q) = \text{MMD}(P,Q) := \sqrt{\E [k(X,X')] - 2 \E[k(X,Y)] + \E[k(Y,Y')]}\]
where $X,X' \sim P$ and $Y,Y' \sim Q$.
\end{enumerate}

In the context of Section~\ref{sec:uq}, we have (at least) the following instantiations of Assumption~\ref{ass:1}:
\begin{enumerate}
    \item[(i)] Assume the reward is bounded by $r_{\max}$. Then (since $\|V^\pi_{\tM}\|_\infty \le \frac{r_{\max}}{1-\gamma}$)
    \[|G^\pi_{\hatM}(s,a)| \le \frac{r_{\max}}{1-\gamma} D_{\textsc{tv}}(\hatT(s,a), \tT(s,a))\]
    This corresponds to $c = \frac{r_{\max}}{1-\gamma}$ and $\mathcal{F} = \{f : \|f\|_\infty \le 1\}$.
    
    \item[(ii)] Assume $V^\pi_{\tM}$ is $L_v$-Lipschitz. Then
    \[|G^\pi_{\hatM}(s,a)| \le L_v W_1(\hatT(s,a), \tT(s,a))\]
    This corresponds to $c = L_v$ and $\mathcal{F} = \{f : \text{$f$ is $1$-Lipschitz}\}$.
    
    \item[(iii)] Assume $\|V^\pi_{\tM}\|_{\mathcal{H}} \le \nu$. Then
    \[|G^\pi_{\hatM}(s,a)| \le \nu \text{MMD}(\hatT(s,a), \tT(s,a))\]
    This corresponds to $c = \nu$ and $\mathcal{F} = \{f : \|f\|_{\mathcal{H}} \le 1\}$.
\end{enumerate}

\section{Proofs}
\label{app:proofs}
We provide a proof for Lemma~\ref{lem:1} for completeness. The proof is essentially the same as that for~\citep[Lemma 4.3]{luo2018algorithmic}.
\begin{proof}
Let $W_j$ be the expected return when executing $\pi$ on $\hatT$ for the first $j$ steps, then switching to $\tT$ for the remainder.
That is,
\[W_j = \Esub{\substack{a_t \sim \pi(s_t) \\ t < j: s_{t+1} \sim \hatT(s_t,a_t) \\ t \ge j: s_{t+1} \sim \tT(s_t,a_t)}}\left[\sum_{t=0}^\infty \gamma^t r(s_t,a_t)\right]\]
Note that $W_0 = \eta_{\tM}(\pi)$ and $W_\infty = \eta_{\hatM}(\pi)$, so
\[\eta_{\hatM}(\pi) - \eta_{\tM}(\pi) = \sum_{j=0}^\infty (W_{j+1}-W_j)\]
Write
\begin{align*}
	W_j &= R_j + \Esub{s_j,a_j \sim \pi, \hatT}\left[\Esub{s_{j+1} \sim \tT(s_t,a_t)}[\gamma^{j+1} V^\pi_{\tM}(s_{j+1})]\right] \\
	W_{j+1} &= R_j + \Esub{s_j,a_j \sim \pi, \hatT}\left[\Esub{s_{j+1} \sim \hatT(s_t,a_t)}[\gamma^{j+1} V^\pi_{\tM}(s_{j+1})]\right]
\end{align*}
where $R_j$ is the expected return of the first $j$ time steps, which are taken with respect to $\hatT$.
Then
\begin{align*}
	W_{j+1}-W_j &= \gamma^{j+1} \Esub{s_j,a_j \sim \pi, \hatT}\left[\Esub{s' \sim \hatT(s_j,a_j)}[V^\pi_{\tM}(s')] - \Esub{s' \sim \tT(s_j,a_j)}[V^\pi_{\tM}(s')]\right] \\
	&= \gamma^{j+1} \Esub{s_j,a_j \sim \pi, \hatT}\left[G^\pi_{\hatM}(s_j,a_j)\right]
\end{align*}
Thus
\begin{align*}
	\eta_{\hatM}(\pi) - \eta_{\tM}(\pi) &= \sum_{j=0}^\infty (W_{j+1}-W_j) \\
	&= \sum_{j=0}^\infty \gamma^{j+1} \Esub{s_j,a_j \sim \pi, \hatT}\left[G^\pi_{\hatM}(s_j,a_j)\right] \\
	&= \gamma \Eisub{(s,a) \sim \rho^\pi_{\hatT}}\left[G^\pi_{\hatM}(s,a)\right]
\end{align*}
as claimed.
\end{proof}

Now we prove Theorem~\ref{thm:main}.
\begin{proof}%
We first note that a two-sided bound follows from Lemma~\ref{lem:1}:
\begin{align}
|\eta_{\hatM}(\pi) - \eta_{\tM}(\pi)| \le \gamma \Eisub{(s,a) \sim \rho^\pi_{\hatT}} |G^\pi_{\hatM}(s,a)| \le \lambda \Eisub{(s,a) \sim \rho^\pi_{\hatT}}[u(s,a)] = \lambda \epsilon_u(\pi) \label{eqn:twosided}
\end{align}
Then we have, for any policy $\pi$,
\begin{align}
\eta_{\tM}(\hat{\pi}) &\ge \eta_{\tilM}(\hat\pi) \tag{by~\eqref{eqn:5}} \\
&\ge \eta_{\tilM}(\pi) \tag{by definition of $\hat\pi$} \\
&= \eta_{\hatM}(\pi) - \lambda \epsilon_u(\pi) \nonumber \\
&\ge \eta_{\tM}(\pi) - 2\lambda\epsilon_u(\pi) \tag{by~\eqref{eqn:twosided}}
\end{align}
\end{proof}

\section{MOPO Practical Algorithm Outline}
\label{app:alg}

We outline the practical MOPO algorithm in Algorithm~\ref{alg:practical}.

\begin{algorithm}[t]
\caption{MOPO instantiation with regularized probabilistic dynamics and ensemble uncertainty}
\small
\label{alg:practical}
	\begin{algorithmic}[1]
\Require reward penalty coefficient $\lambda$ rollout horizon $h$, rollout batch size $b$. 
\State Train on batch data $\mathcal{D}_{\text{env}}$ an ensemble of $N$ probabilistic dynamics $\{\hatT^i(s', r \given s, a) = \mathcal{N}(\mu^i(s, a), \Sigma^i(s, a))\}_{i=1}^N$.
		\State Initialize policy $\pi$ and empty replay buffer $\mathcal{D}_{\text{model}} \leftarrow \varnothing$.
		\For{epoch $1, 2, \dots$} \Comment{This for-loop is essentially one outer iteration of MBPO} \label{line:1}
		    \For{$1, 2, \dots, b$ (in parallel)}
		        \State Sample state $s_1$ from $\mathcal{D}_{\text{env}}$ for the initialization of the rollout.
		        \For{$j = 1, 2, \dots, h$}
		            \State Sample an action $a_j \sim \pi(s_j)$.
		            \State Randomly pick dynamics $\hatT$ from $\{\hatT^i\}_{i=1}^N$ and sample $s_{j+1},r_j \sim \hatT(s_j,a_j)$.
		            \State Compute $\tilde{r}_j = r_j {- \lambda \max_{i=1}^N\|\Sigma^i(s_j,a_j)\|_\text{F}}$.
		            \State Add sample $(s_j, a_j, \tilde{r}_j, s_{j+1})$ to $\mathcal{D}_{\text{model}}$. \label{line:2}
		        \EndFor
		    \EndFor
		    \State Drawing samples from $\mathcal{D}_{\text{env}} \cup \mathcal{D}_{\text{model}}$, use SAC to update $\pi$. 
		\EndFor
	\end{algorithmic}
\end{algorithm}

\section{Ablation Study}
\label{app:ablation}
To answer question (3), we conduct a thorough ablation study on MOPO. The main goal of the ablation study is to understand how the choice of reward penalty affects performance. We denote \textbf{no ens.} as a method without model ensembles, \textbf{ens. pen.} as a method that uses model ensemble disagreement as the reward penalty, \textbf{no pen.} as a method without reward penalty, and \textbf{true pen.} as a method using the true model prediction error $\|\hatT(s,a) - \tT(s,a)\|$ as the reward penalty.
Note that we include \textbf{true pen.} to indicate the upper bound of our approach. Also, note that \textbf{no ens.} measures disagreement among the ensemble: precisely, if the models’ mean predictions are denoted $\mu_1,\dots, \mu_N$, we compute the average $\bar\mu = 1/N \sum_{i=1}^N \mu_i$ and then take $\max_i \|\mu_i - \bar\mu\|$ as the ensemble penalty.

The results of our study are shown in Table~\ref{fig:ablation}.
For different reward penalty types, reward penalties based on learned variance perform comparably to those based on ensemble disagreement in D4RL environments while outperforming those based on ensemble disagreement in out-of-distribution domains. Both reward penalties achieve significantly better performances than no reward penalty, indicating that it is imperative to consider model uncertainty in batch model-based RL.
Methods that uses oracle uncertainty obtain slightly better performance than most of our methods. Note that \textbf{MOPO} even attains the best results on \texttt{halfcheetah-jump}. Such results suggest that our uncertainty quantification on states is empirically successful, %
since there is only a small gap. We believe future work on improving uncertainty estimation may be able to bridge this gap further. Note that we do not report the results of methods with oracle uncertainty on \texttt{walker2d-mixed} and \texttt{ant-angle} as we are not able to get the true model error from the simulator based on the pre-recorded dataset.

In general, we find that performance differences are much larger for \texttt{halfcheetah-jump} and \texttt{ant-angle} than the D4RL \texttt{halfcheetah-mixed} and \texttt{walker2d-mixed} datasets, likely because \texttt{halfcheetah-jump} and \texttt{ant-angle} requires greater generalization and hence places more demands on the accuracy of the model and uncertainty estimate.

Finally, we perform another ablation study on the choice of the reward penalty. We consider the $u^\text{mean}(s,a) = \frac{1}{N}\sum_{i=1}^N\|\Sigma^i_\phi(s, a)\|_\text{F}$, the average standard deviation of the learned models in the ensemble, as the reward penalty instead of the max standard deviation as used in MOPO. We denote the variant of MOPO with the average learned standard deviation as \textbf{MOPO, avg. var.}. We compare \textbf{MOPO} to \textbf{MOPO, avg. var.} in the halfcheetah-jump domain. \textbf{MOPO} achieves \textbf{4140.6}$\pm$88 average return while \textbf{MOPO, avg. var.} achieves \textbf{4166.3}$\pm$228.8 where the results are averaged over 3 random seeds. The two methods did similarly, suggesting that using either mean variance or max variance would be a reasonable choice for penalizing uncertainty.

\begin{table}[h]
\centering
\small
\begin{tabular}{l|r|r|r|r}
\toprule
\textbf{Method} & halfcheetah-mixed & walker2d-mixed & halfcheetah-jump & ant-angle\\ \midrule
\textbf{MOPO} & $6405.8\pm35$ & $1916.4\pm611$ & $4016.6\pm144$ & $2530.9\pm137$\\
\textbf{MOPO, ens. pen.} & $6448.7\pm115$ & $1923.6\pm752$ & $3577.3\pm461$ & $2256.0 \pm 288$\\
\textbf{MOPO, no pen.} & $6409.1\pm429$ & $1421.2\pm359$ & $-980.8\pm5625$ & $18.6\pm49$\\
\textbf{MBPO} & $5598.4\pm1285$ & $1021.8 \pm 586$ & $2971.4\pm1262$ & $13.6\pm65$\\
\textbf{MBPO, no ens.} & $2247.2\pm581$ & $500.3\pm34$ & $-68.7\pm1936$ & $-720.1\pm728$\\
\midrule
\textbf{MOPO, true pen.} & $6984.0\pm148$ & N/A & $3818.6\pm136$  & N/A\\
\bottomrule
\end{tabular}
\vspace{0.1cm}
\caption{\footnotesize Ablation study on two D4RL tasks \texttt{halfcheetah-mixed} and \texttt{walker2d-mixed} and two out-of-distribution tasks \texttt{halfcheetah-jump} and \texttt{ant-angle}. We use average returns where the results of MOPO and its variants are averaged over 6 random seeds and MBPO results are averaged over 3 random seeds as in Table~\ref{fig:generalize}. We observe that different reward penalties can all lead to substantial improvement of the performance and reward penalty based on learned variance is a better choice than that based on ensemble disagreement in out-of-distribution cases. Methods that use oracle uncertainty as the reward penalty achieve marginally better performance than \textbf{MOPO}, implying that \textbf{MOPO} is effective at estimating the uncertainty. }
\label{fig:ablation}
\vspace{-0.6cm}
\normalsize
\end{table}

\section{Empirical results on generalization capabilities}
\label{app:generalization_limit}
We conduct experiments in \textit{ant-angle} to show the limit of MOPO's generalization capabilties. As shown in Table~\ref{tab:ant}, we show that MOPO generalizes to Ant running at a $45^\circ$ angle (achieving almost buffer max score), beyond the $30^\circ$ shown in the paper, while failing to generalize to a $60$ and $90^\circ$ degree angle.
This suggests that if the new task requires to explore states that are completely out of the data support, i.e. the buffer max and buffer mean both fairly bad, MOPO is unable to generalize.

\begin{table}[h]
\centering
\footnotesize
\begin{tabular}{cccc}
\toprule
\textbf{Environment} & \textbf{Buffer Max} & \textbf{Buffer Mean} & \textbf{MOPO}\\ \hline
\texttt{ant-angle-45} & 3168.7 & 1105.5 & 2571.3$\pm$598.1\\
\texttt{ant-angle-60} & 1953.7 & 846.7 & 840.5$\pm$1103.7\\
\texttt{ant-angle-90} & 838.8 & -901.6 & -503.2$\pm$803.4\\
\bottomrule
\end{tabular}
\caption{\footnotesize Limit of generalization on \texttt{ant-angle}.}
\label{tab:ant}
\normalsize
\end{table}

\section{Experiments on HIV domains}
\label{app:hiv}

Beyond continous control tasks in MuJoCo, we test  MOPO on an HIV treatment simulator slightly modified from the one in the \href{https://github.com/zykls/whynot}{whynot package}. The task simulates the sequential decision making in HIV treatment, which involves determining the amounts of two anti-HIV drugs to be administered to the patient in order to maximize the immune response and minimize the amount of virus. The agent observes both of those quantities as well as the (log) number of infected and uninfected T cells and macrophages.

We evaluated MOPO with the data generated from the first 200k steps of training an online SAC agent on this environment.
We show results in Table~\ref{tab:hiv}, where MOPO outperforms BEAR and achieves almost the buffer max score.
\begin{table}[h]
\centering
\footnotesize
\begin{tabular}{cccccc}
\toprule
\textbf{Buffer Max} & \textbf{Buffer Mean} & \textbf{SAC (online)} & \textbf{BEAR} & \textbf{MOPO}\\ \hline
15986.2 & 6747.2 & $25716.3 \pm 254.3$ & 11709.1$\pm$1292.1 & $\mathbf{13484.6}\pm3900.7$\\
\bottomrule
\end{tabular}
\caption{\footnotesize HIV treatment results, averaged over 3 random seeds.}
\label{tab:hiv}
\normalsize
\end{table}

\section{Experiment Details}
\label{app:details}

\subsection{Details of out-of-distribution environments}
For \texttt{halfcheetah-jump}, the reward function that we use to train the behavioral policy is $r(s, a) = \max\{v_x, 3\} - 0.1*\|a\|_2^2$ where $v_x$ denotes the velocity along the x-axis. After collecting the offline dataset, we relabel the reward function to $r(s, a) = \max\{v_x, 3\} - 0.1*\|a\|_2^2 + 15*(z - \text{init z})$ where $z$ denotes the z-position of the half-cheetah and $\text{init z}$ denotes the initial z-position.

For \texttt{ant-angle}, the reward function that we use to train the behavioral policy is $r(s, a) = v_x - \text{control cost}$. After collecting the offline dataset, we relabel the reward function to $r(s, a) = v_x\cdot\cos\frac{\pi}{6} + v_y\cdot\sin\frac{\pi}{6} - \text{control cost}$ where $v_x$, $v_y$ denote the velocity along the $x, y$-axis respectively.

For both out-of-distribution environments, instead of sampling actions from the learned policy during the model rollout (line 10 in Algorithm~\ref{alg:practical}), we sample random actions from $\text{Unif}[-1, 1]$, which achieves better performance empirically. One potential reason is that using random actions during model rollouts leads to better exploration of the OOD states. 

\subsection{Hyperparameters}
\label{app:hyperparameters}
Here we list the hyperparameters used in the experiments.

For the D4RL datasets, the rollout length $h$ and penalty coefficient $\lambda$ are given in Table~\ref{fig:d4rl-hyperparams}.
We search over $(h,\lambda) \in \{1,5\}^2$ and report the best final performance, averaged over 3 seeds.
The only exceptions are halfcheetah-random and walker2d-medium-expert, where other penalty coefficients were found to work better.

\begin{table}
\centering
\small
\begin{tabular}{l|l|c|c}
\toprule
\textbf{\!\!\!Dataset type\!\!} & \textbf{Environment} & MOPO $(h, \lambda)$ & MBPO $h$ \\ \midrule
\!\!\!random & halfcheetah & 5, 0.5 & 5 \\
\!\!\!random & hopper & 5, 1 & 5 \\
\!\!\!random & walker2d & 1, 1 & 5  \\
\!\!\!medium & halfcheetah & 1, 1 &  5 \\
\!\!\!medium & hopper & 5, 5 & 5  \\
\!\!\!medium & walker2d & 5, 5 & 5 \\
\!\!\!mixed & halfcheetah & 5, 1 & 5 \\
\!\!\!mixed & hopper & 5, 1 & 5 \\
\!\!\!mixed & walker2d & 1, 1 & 1 \\
\!\!\!med-expert\!\!\! & halfcheetah & 5, 1 &  5 \\
\!\!\!med-expert\!\!\! & hopper & 5, 1 & 5 \\
\!\!\!med-expert\!\!\! & walker2d & 1, 2 & 1 \\
\bottomrule
\end{tabular}
\caption{Hyperparameters used in the D4RL datasets.}
\label{fig:d4rl-hyperparams}
\normalsize
\end{table}

For the out-of-generalization tasks, we use rollout length $5$ for \texttt{halfcheetah-jump} and $25$ for \texttt{ant-angle}, and penalty coefficient $1$ for \texttt{halfcheetah-jump} and $2$ for \texttt{ant-angle}.

Across all domains, we train an ensemble of $7$ models and pick the best $5$ models based on their prediction error on a hold-out set of $1000$ transitions in the offline dataset. Each of the model in the ensemble is parametrized as a 4-layer feedforward neural network with $200$ hidden units and after the last hidden layer, the model outputs the mean and variance using a two-head architecture. Spectral normalization~\cite{miyato2018spectral} is applied to all layers except the head that outputs the model variance.

For the SAC updates, we sample a batch of $256$ transitions, $5\%$ of them from $\Denv$ and the rest of them from $\Dmodel$. We also perform ablation studies on the percentage of the real data in a batch for MOPO. For simplicity, we use MBPO, which essentially MOPO without reward penalty, for this ablation study. We tried to train MBPO with data all sampled from $\Dmodel$ and no data from $\Denv$ and compare the performance to MBPO with $5\%$ of data from $\Denv$ on all 12 settings in the D4RL benchmark. We find that the performances of both methods are not significantly distinct: no-real-data MBPO outperforms $5\%$-real-data MBPO on 6 out of 12 tasks and lies within one SD of $5\%$-real-data MBPO on 9 out of 12 tasks.

\end{document}